%% file: main.tex
\documentclass[final,12pt]{elsarticle}

\usepackage{amssymb}
\usepackage{amsmath}
\usepackage{float}
\usepackage{pifont}
\usepackage{todonotes}
\usepackage{multirow}
\usepackage{array}
\usepackage{stfloats}
\usepackage{placeins}
\usepackage{comment}
\usepackage[misc]{ifsym}

\definecolor{ves_col}{HTML}{803937}
\definecolor{ica_col}{HTML}{6C8EBF}
\definecolor{vesica_col}{HTML}{D79B00}
\definecolor{fica_col}{HTML}{9673A6}
\definecolor{fves_col}{HTML}{82B366}
\definecolor{ves_spatial}{HTML}{CC0000}
\definecolor{ica_spatial}{HTML}{0075E8}
\definecolor{faz_spatial}{HTML}{007E01}

\begin{document}

\begin{frontmatter}



\title{Interpretable Retinal Disease Prediction Using Biology-Informed Heterogeneous Graph Representations}


\author{Laurin Lux \Letter \ \textit{laurin.lux@tum.de}, Alexander H. Berger, Maria Romeo Tricas, Richard Rosen, Alaa E. Fayed, Sobha Sivaprasada, Linus Kreitner, Jonas Weidner, Martin J. Menten, Daniel Rueckert and Johannes C. Paetzold} 

\affiliation{organization={Munich Center for Machine Learning (MCML)},
            city={Munich},
            country={Germany}}

\affiliation{organization={School of Medicine and Health, TUM University Hospital},
            city={Munich},
            country={Germany}}

\affiliation{organization={School of Computation, Information and Technology, TUM},
            city={Munich},
            country={Germany}}

\affiliation{organization={Weill Cornell Medicine, Cornell University, New York City},
            city={New York},
            country={USA}}


\begin{abstract}
Interpretability is crucial for utilizing machine learning models as clinical decision support tools for medical diagnostics. However, most state-of-the-art image classifiers based on neural networks are not interpretable. As a result, clinicians often resort to known biomarkers to guide diagnosis, although biomarker-based classification often suffers from drastic information loss compared to raw medical images. This work proposes a method that preserves the rich imaging information while simultaneously enhancing the interpretability of predictions for diabetic retinopathy staging from optical coherence tomography angiography (OCTA) images. 
The core contribution of our method is a novel biology-informed heterogeneous graph representation that models retinal vessel segments, intercapillary areas, and the foveal avascular zone (FAZ) in a human-interpretable way. This graph representation allows us to frame diabetic retinopathy staging as a graph-level classification task, which we solve using an established, efficient graph neural network architecture.
We compare our method against established methods, including classical biomarker-based classifiers, convolutional neural networks (CNNs), and vision transformers in predicting the clinically assigned DR stage based on color fundus photography images. We find stage agreement rates of our method and alternative vision model based classifiers saturating at AUC-ROC values of 84\%. Crucially, we use our biology-informed graph to provide explanations of great detail. Our approach surpasses existing methods in precisely localizing and identifying abnormal vessels and non-perfusion areas. Our approach sets the stage for the interpretable identification of patients who require special attention due to their traceable microvascular changes, only observable using the details of OCTA images. 
\end{abstract}

\end{frontmatter}



\section{Introduction}
\label{sec:introduction}
Diabetic Retinopathy (DR), a complication of diabetes that affects the retinal vasculature, is one of the leading causes of blindness in adulthood \cite{teo2021global}. It is associated with pathological changes to the retinal microvasculature, such as loss of capillaries through capillary degeneration, the occurrence of intraretinal microvascular abnormalities, microaneurysms, and also neovascularization. 

Optical coherence tomography angiography (OCTA), a non-invasive and high-resolution retinal imaging technique \cite{le2021machine}, has emerged as a powerful tool for studying these microvascular changes. OCTA enables detailed examination of the retinal vasculature \cite{fayed2024retinal} down to the capillary level by resolving the superficial vascular complex (SVC) and deep vascular complex (DVC), facilitating early-stage DR detection \cite{moore1999three}.

Currently, clinicians study quantitative biomarkers extracted from OCTA images to characterize DR-related microvascular alterations, including vessel density, vessel tortuosity, and foveal avascular zone (FAZ) area. However, these biomarkers have important limitations: they fail to fully capture complex pathological changes such as intraretinal microvascular abnormalities and neovascularization, and simple metrics like tortuosity index and vessel density lack spatial localization, making clinical verification challenging. Deep learning models, that are already well-established for other imaging modalities \cite{stolte2020survey}, offer the potential to extract the full spectrum of microvascular pathological information from OCTA images to guide clinical diagnosis beyond traditional biomarkers. Such AI-based OCTA analysis could serve as a valuable support tool within the multimodal diagnostic workflow that integrates OCT, color fundus photography, fluorescein angiography, and patient history. OCTA-supported DR staging is not yet a clinical standard procedure in the diagnostic workflow. However, numerous studies have demonstrated the diagnostic prowess of OCTA images, which are highly valuable for a detailed disease characterization. OCTA image analysis enables early detection of signs of DR in patients suffering from diabetes \cite{de2015detection,takase2015enlargement,sun2021optical}. Moreover, OCTA scans provide predictive information about the risk of progression to a vision-threatening form of DR 
\cite{sun2019oct,ding2025expanded,marques2024patterns,yang2023assessment}. In particular, Sun et al.~\cite{sun2019oct} demonstrated prospectively in 205 eyes that OCTA-derived metrics of the deep capillary plexus predicted DR progression and diabetic macular edema development beyond established clinical risk factors. The integration of OCTA imaging as a supportive modality into the existing clinical workflow is, therefore, a promising addition for more targeted follow-up and treatment decisions. Tailoring screening intervals to progression risk can help drastically reduce the burden on clinical experts and use resources where they are needed most. With an estimated 9.6 million people affected by DR in the US alone, the potential impact of such risk stratification on healthcare resource allocation is substantial.

\begin{figure}[t]
    \centering
    \includegraphics[trim={0cm 0cm 1cm 0cm},clip,width = \linewidth]{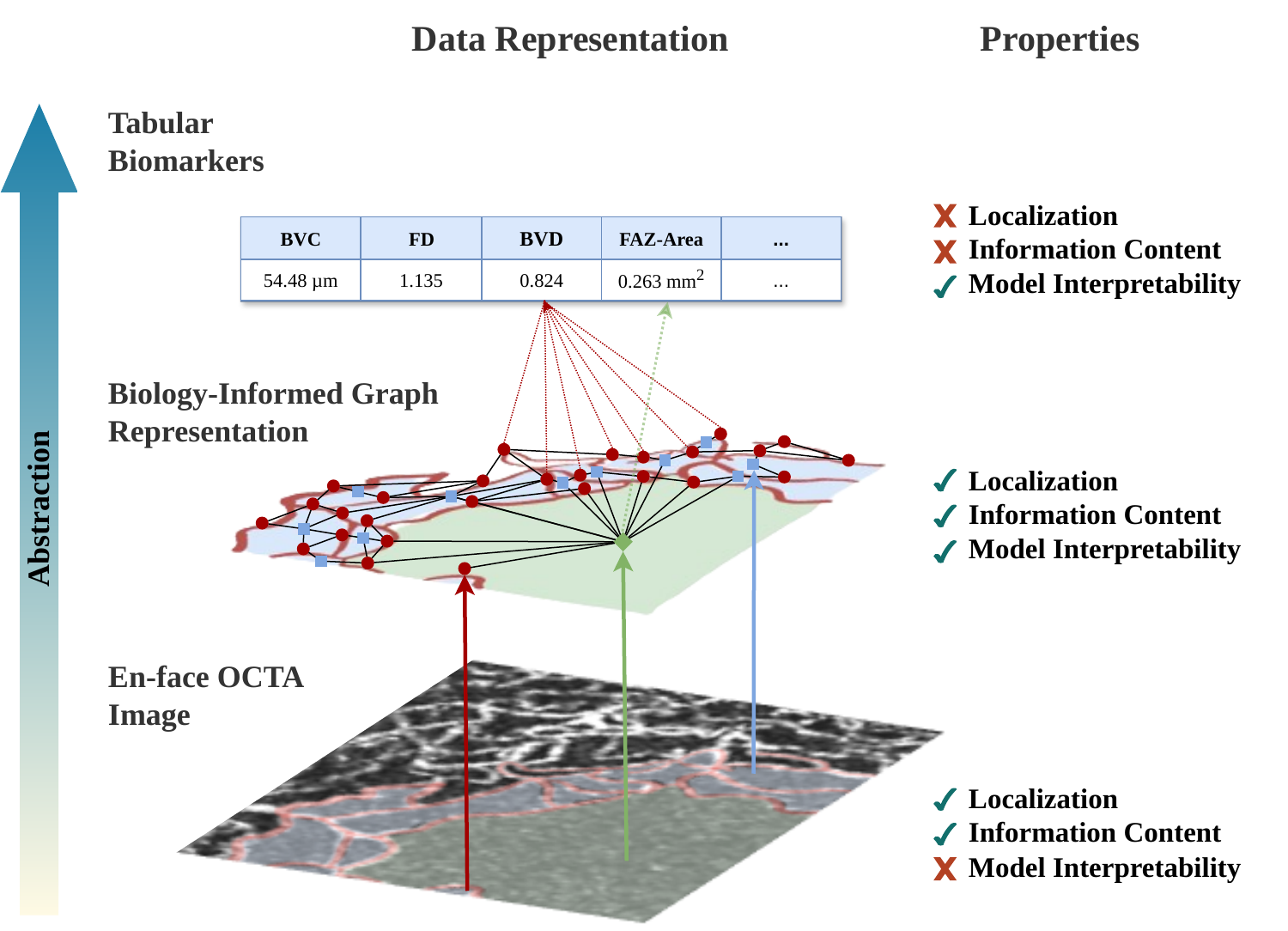}
    \caption{Schematic illustration of our heterogeneous graph representing the retinal vasculature. We elevate the image to a higher abstraction level, where nodes represent understandable biological concepts, such as vessels, intercapillary areas, and the FAZ. Because of each node's biological meaning, known tabular biomarkers are naturally encoded in our heterogeneous graph. Consequently, classifiers trained on our representation combine the favorable interpretability of tabular biomarkers with image-level localization and performance.}
    \vspace{-0.5cm}
    \label{fig:abstraction}
\end{figure}

Current Machine Learning (ML)-based approaches trained on OCTA images and OCTA-extracted biomarkers have shown good performance for diagnosing DR \cite{le2021machine, alam2019supervised, sandhu2020automated, alam2020quantitative, heisler2020ensemble, ryu2021deep, zang2022diabetic}. However, these deep learning (DL) approaches are not used for clinical decision-making, partially because their black-box predictions cannot be effectively combined with the multi-modal information sources that ophthalmologists routinely use for DR assessment. Specifically, the lack of interpretability makes it difficult for clinicians to understand how OCTA-based predictions relate to findings from other imaging modalities and patient factors, hindering their integration into comprehensive diagnostic evaluations. Although DL models may implicitly leverage high-level pathological concepts (e.g., non-perfusion areas or microaneurysms), existing explainability methods fail to attribute the model's decisions to these concepts. For convolutional neural networks (CNNs), CAM \cite{zhou2016learning} and grad-CAM \cite{selvaraju2017grad} are common methods that focus on localizing critical image regions. Previous studies \cite{heisler2020ensemble, ryu2021deep, zang2022diabetic} employed these explainability methods for DR classification, yielding mixed results. While their approaches allowed a rough identification of critical image regions, the emphasized areas remain broad and unspecific. Moreover, the most critical shortcoming of these methods is the semantic gap between the highlighted image regions and human-interpretable concepts. Unlike DL approaches, inherently interpretable models, such as simple decision trees based on human-understandable biomarkers, do not achieve state-of-the-art prediction accuracy.

To address the challenge of interpretable OCTA image analysis with DL, we introduce a novel graph representation for OCTA images for accurate and interpretable DR staging.\footnote{Code available at https://github.com/luxtu/OCTA-graph.}  The main novelty of this work lies in this biology-informed graph representation rather than in the learning architecture or the attribution method: both are established techniques that we adapt to our representation to demonstrate its utility. Specifically, our contributions are as follows:

\begin{enumerate}
    \item We introduce a heterogeneous graph representation that expresses the biological structures in OCTA images. It preserves spatial and semantic information content while abstracting from the raw image representation to make it interpretable by clinicians. Furthermore, this heterogeneous graph representation preserves neighborhood information by incorporating both homogeneous and heterogeneous edges. Figure \ref{fig:abstraction} illustrates how our approach combines the raw image's precise spatial information with the interpretative power of high-level biomarkers. The proposed representation is a foundation for diverse classification or correlation tasks, utilizing convolutional architectures such as graph neural networks (GNNs).

    \item We apply an established heterogeneous message-passing GNN architecture and an established attribution method (integrated gradients) to our new representation for DR staging. This combination allows us to directly link stage predictions to clinically well-studied indicators of DR, such as abnormal vessel segments, microaneurysms, and retinal areas with decreased capillary density. With this application, we highlight the potential of our graph-based representation for further downstream tasks; the interpretability gains stem from the representation, not from the learning or attribution components themselves.

    \item Finally, we utilize our interpretable model explanations combined with OCTA expert readings to identify a subgroup of patients where microvascular pathologies observed in 3 mm x 3 mm OCTA images deviate from the clinical DR stage based on color fundus images. This enables the identification of patients who require special attention for follow-up and treatment decisions.

\end{enumerate}

By providing an interpretable data representation combined with an explainable and high-performing prediction algorithm, we aim to advance support tools for clinical decision-making.

\section{Related Work}
\label{sec:related_work}
\begin{figure*}[!ht]
    \centering
    \includegraphics[
    width = \textwidth]{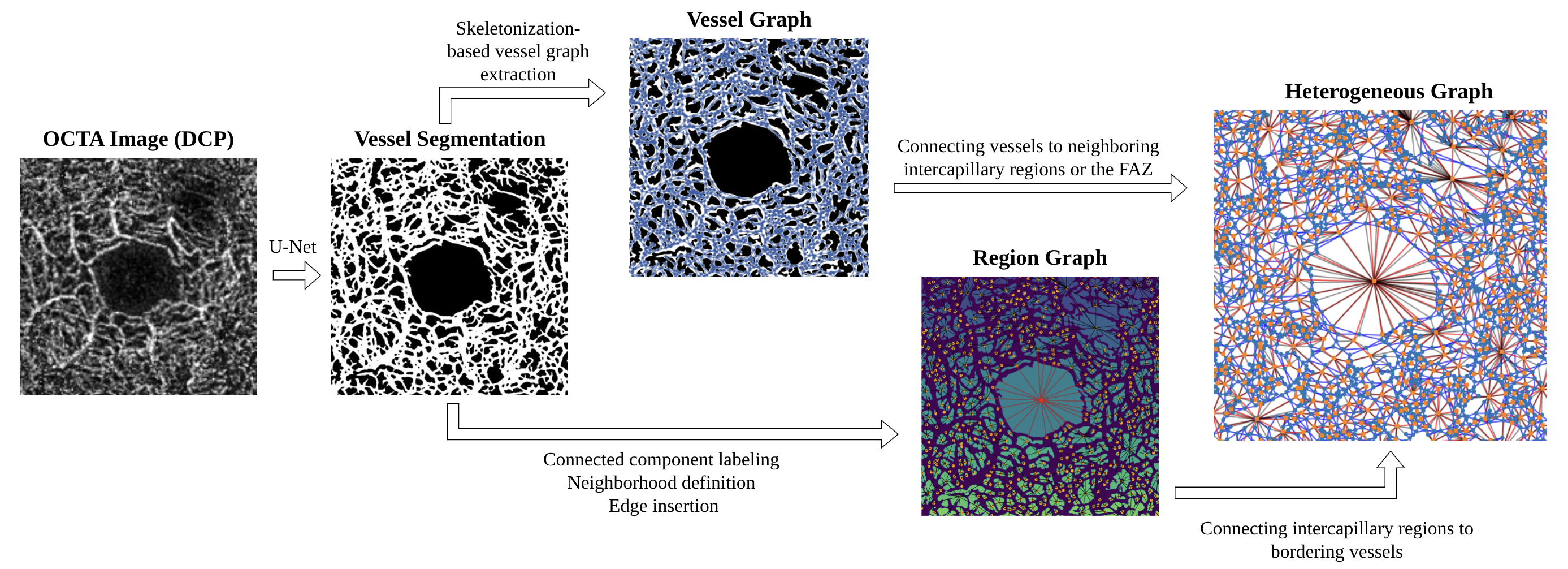}
    \caption{Processing pipeline for generating a heterogeneous graph that models an OCTA image's most relevant biological concepts: vessels, intercapillary areas, and the FAZ. A vasculature segmentation \cite{kreitner2023detailed} is used to create a vessel graph and intercapillary area graph and identify the FAZ. Finally, these components are merged into a single interconnected heterogeneous graph representation.}
        \vspace{-0.3cm}
    \label{fig:construction}
\end{figure*}

\subsection{Machine Learning for DR Staging in OCTA} 
In recent years, the improvement of instrumentation \cite{kim2011vivo} has made OCTA available to an increasing number of ophthalmologists. Since then, many studies have investigated DR with OCTA. Approaches to staging DR with OCTA can be broadly categorized into two groups. 
The first group comprises feature-based methods, which are based on prior knowledge of indicative biomarkers, the quantitative assessment of these markers, and decision-making based on the extracted information \cite{le2021machine, alam2019supervised, sandhu2020automated, alam2020quantitative}. Traditionally utilized biomarkers include blood vessel density (BVD), blood vessel caliber (BVC), blood vessel tortuosity, vessel perimeter index, vessel complexity index, FAZ area, and FAZ contour irregularity. These biomarkers serve as inputs for classification models such as neural networks, support vector machines (SVM), or tree-based methods such as random forests (RF). Despite their widespread acceptance and proven significance for DR disease staging, feature-based approaches have limitations. They can only rely on a priori defined biomarkers, established based on human expertise and laborious verification procedures. To date, no existing biomarker quantifies the severity of intraretinal microvascular abnormalities as well as the severity of microaneurysms in OCTA images. Moreover, in the classification setting, existing biomarkers are usually evaluated as global metrics without localization information.

The second group comprises DL models directly operating on the raw OCTA images. Previous studies adopted well-established CNN architectures such as ResNet \cite{xie2017aggregated}, VGG \cite{simonyan2014very}, and EfficientNet \cite{tan2019efficientnet} for DR staging and achieved high accuracy. Zhou et al. proposed a transformer-based foundation model for retinal images \cite{zhou2023foundation}, which outperformed other state-of-the-art approaches for DR staging on color fundus images. Although the network was not directly trained on OCTA images, the model has a detailed understanding of abstract retinal concepts by training on a large amount of data.
DL-based methods have shown favorable performance compared to the aforementioned traditional approaches. This performance gap indicates that DL methods can extract information that, to date, has not been formalized as an established, known biomarker, such as the severity and frequency of intraretinal microvascular abnormalities. They can be combined with explainability tools, such as CAM \cite{zhou2016learning} and grad-CAM \cite{selvaraju2017grad}, to highlight critical image regions \cite{heisler2020ensemble, ryu2021deep, zang2022diabetic}. However, the human-in-the-loop cannot overcome the semantic gap between highlighted image regions and their corresponding intensity values and the abstract concepts established later in the CNN network. Clinicians cannot verify and interpret the algorithmic outcomes beyond the localization of influential regions. Moreover, the biomarkers utilized by the CNN networks cannot be identified and formalized.

\subsection{Graph Neural Networks as Explainable Models} 
With the advent of graph neural networks (GNNs), convolutions were generalized towards graph-structured data; oftentimes including the concept of message-passing \cite{velickovic2017graph, kipf2016semi, hamilton2017inductive}. 
This has led to the widespread application of GNNs across diverse fields, including chemistry, physics, medicine, and social sciences, where graph-structured data is abundant \cite{zhou2020graph}. In biomedical image analysis, successful applications of GNNs to discriminative tasks include brain disorder classification from functional magnetic resonance images \cite{zhang2022classification}, multiple sclerosis disease activity prediction from MR images \cite{prabhakar2023self}, vessel hierarchy classification in microscopic images \cite{paetzold2021whole}, and pathology image classification \cite{wang2023ccf}. 

In tandem with the evolution and utilization of GNN architectures, the methods for explaining their predictions have advanced. For example, gradient-based approaches, proven successful in computer vision and CNNs, effectively provided explanations for graph-based molecular property prediction \cite{pope2019explainability}. Specific perturbation-based techniques tailored for graph-structured data, such as GNNExplainer \cite{ying2019gnnexplainer}, have been developed. Notably, GraphLime \cite{huang2022graphlime}, drawing inspiration from LIME \cite{ribeiro2016should}, stands out as a surrogate-based explainability method that uses local perturbations. For a comprehensive overview of other gradient-, perturbation-, or surrogate-based explanation approaches, refer to \cite{yuan2022explainability}.
Agarwal et al. \cite{agarwal2023evaluating} analyzed various explanation methods tailored for graph classification tasks. Their study showed that integrated gradients is the most faithful and accurate explanation method, particularly on real-world datasets with known ground truth explanations.

\subsection{Advances in Interpretability of Vision Models}
CNNs and Vision Transformers have revolutionized medical computer vision by achieving state-of-the-art performance across essential tasks such as medical image segmentation and classification. The interpretability of these models has lagged behind their impressive accuracy. Post-hoc explainability methods have, until now, been the gold standard for interpretability in medical vision models. With the most prominent approaches being  CAM \cite{zhou2016learning}, Grad-CAM \cite{selvaraju2017grad}, Integrated Gradients \cite{sundararajan2017axiomatic}, Saliency \cite{simonyan2013deep} and LIME \cite{ribeiro2016should}. Numerous other methods specifically targeted to medical applications have been proposed \cite{van2022explainable}. Recent developments aim to overcome the shortcomings shared by all of these methods; pixel-wise intensity attribution does not provide insights into high-level concepts, which is particularly critical for high-stakes decisions such as medical diagnosis. Recent work on mechanistic interpretability seeks to overcome this limitation by decomposing models' internal representations into human-understandable monosemantic features \cite{pach2025sparse}.

\section{Methods}
\label{sec:methods}
\subsection{Construction of a Heterogeneous Graph Representation}

We establish a heterogeneous graph representation for OCTA images incorporating biological domain knowledge. This representation models biological elements found in retinal images, specifically vessel segments, intercapillary areas, and the FAZ. Moreover, we preserve neighborhood information through edge connections, which are subsequently leveraged by graph convolutional algorithms. The process of constructing the heterogeneous graph is depicted in Figure \ref{fig:construction}. Figure \ref{fig:schematic} provides a detailed illustration of the graph's edges and nodes.

\subsubsection{Vessel Segmentation}
We generate vessel segmentation maps using the method by Kreitner et al. \cite{kreitner2023detailed, menten2023synthetic}. This method leverages synthetic training data to create high-fidelity label maps of the retinal vasculature. The resulting label maps include even the smallest capillary vessels and are highly continuous, which is important to generate continuous vessel and intercapillary area graphs.

\subsubsection{Vessel Graph}
Our representation incorporates vessel graphs \cite{paetzold2023geometric, todorov2020machine} to model the vasculature in imaged eyes. Biomarkers such as vessel complexity and diameter are also encapsulated in our vessel graphs. Each vessel segment between two bifurcation points is designated as a node, and edges between vessel nodes are established using bifurcation points. The vasculature segmentation serves as input for a graph extraction algorithm \cite{drees2021scalable}, which extracts the vessel graph and provides feature descriptors detailing properties such as volume, length, curvature, and surface-distance-based values for the centerline. These features serve as embeddings for the nodes in the vessel graph. For detailed descriptions of the extracted geometric features, refer to \cite{drees2021scalable} and our GitHub repository. Figure \ref{fig:schematic} depicts the vessel segments as nodes $v_{ves}$ and the edges between adjacent vessel segments as $e_{ves}$. We use the open-source Voreen framework, built to visualize and explore scientific data to run the graph extraction algorithm \cite{meyer2009voreen}.

\subsubsection{Intercapillary Area Graph}
Intercapillary areas are indicators for disease progression in DR \cite{schottenhamml2016automatic, terada2022intercapillary}. Essentially, they are the counterparts of segmented vessels in OCTA images. We create the nodes of the intercapillary area graph using connected component labeling of the inverted segmentation map. The nodes of the intercapillary area graph are enriched with geometric descriptors such as total area, perimeter length, and eccentricity. We rely on the skeleton of the vessel segmentation to establish the edges in the intercapillary area graph. In the skeletonized segmentation map, we find edge candidates by traversing all skeleton pixels and checking if directly adjacent pixels are part of separate background components of the skeletonized segmentation map. Using the injective map of background components in the segmentation map to background components in the skeletonized segmentation map, edges are then introduced between the intercapillary areas. In Figure \ref{fig:schematic}, the intercapillary area nodes $v_{\scriptscriptstyle ICA}$ represent the green areas (blue nodes), and neighborhood information is encoded in the $e_{\scriptscriptstyle ICA}$ edge type.

\subsubsection{Foveal avascular zone} 
\begin{figure}[!t]
    \centering
    \includegraphics[trim={8.5cm 4.8cm 7.65cm 5.3cm},clip,width = 0.90\linewidth]
    {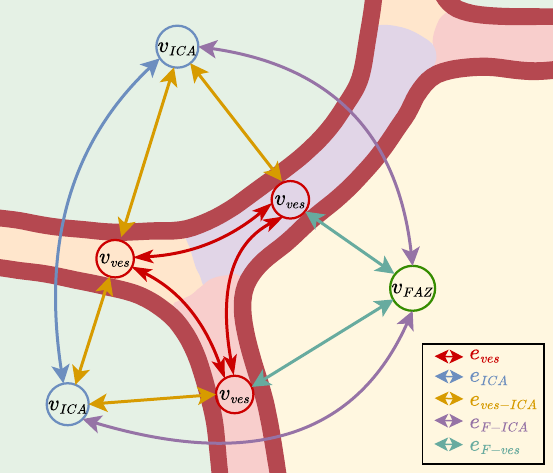}
    \caption{Schematic illustration of our heterogenous graph representation of an OCTA image. Raw pixels in the image are abstracted into either a vessel node ($v_{\scriptscriptstyle ves}$), an intercapillary area node ($v_{\scriptscriptstyle ICA}$), or the FAZ ($v_{\scriptscriptstyle FAZ}$). Neighborhood information is preserved in the homogeneous (\textcolor{ves_col}{$e_{\scriptscriptstyle ves}$}, \textcolor{ica_col}{$e_{\scriptscriptstyle ICA}$}) and heterogeneous edges (\textcolor{vesica_col}{$e_{\scriptscriptstyle ves-ICA}$}, \textcolor{fica_col}{$e_{\scriptscriptstyle F-ICA}$}, \textcolor{fves_col}{$e_{\scriptscriptstyle F-ves}$)}. Interpretable geometric descriptors and intensity statistics are encoded in the node embeddings. }
    \vspace{-0.3cm}
    \label{fig:schematic}
\end{figure}

Because of its significance as a biomarker \cite{sun2021optical, conrath2005foveal}, we assign a separate node type and unique edge types to the FAZ. The FAZ is identified as the central intercapillary area of the connected component labeling. Our used datasets are FAZ-centered during preprocessing; note that this is a necessary prerequisite for correctly identifying the FAZ. The separate node type for the FAZ adds an inductive bias for downstream algorithms. The FAZ is represented through the node $v_{\scriptscriptstyle FAZ}$ (see Figure \ref{fig:schematic}).

\subsubsection{Heterogeneous Edges}

In addition to the diverse node types, our representation includes heterogeneous edge connections between nodes of different types. Edges link intercapillary areas and vessels ($e_{\scriptscriptstyle ves-ICA}$) if an edge forms part of the border of an intercapillary area. For the FAZ, edges connect to vessels along its border ($e_{\scriptscriptstyle F-ves}$). Edges from the FAZ to intercapillary areas are created when a single vessel acts as a bisector to an intercapillary area ($e_{\scriptscriptstyle F-ICA}$). \\

Collectively, this biology-inspired graph representation introduces bias for downstream tasks while enhancing human interpretability by abstracting from gray values to intuitive concepts like vessels. Additionally, the graph representation preserves spatial information through superpixels in the image and neighborhood information through edges. Positional coordinates can be embedded via x- and y-coordinates for intercapillary areas and FAZ, as well as x- and y-coordinates for the start and endpoints of vessel segments. Our graph representation can be used as a foundation for various prominent prediction tasks on OCTA images (see Section \ref{sec:method:staging} and \ref{sec:discussion}). 

\subsection{Graph Learning for DR Staging}
\label{sec:method:staging}
\begin{figure*}[!ht]
    \centering
    \includegraphics[
    width = \textwidth]{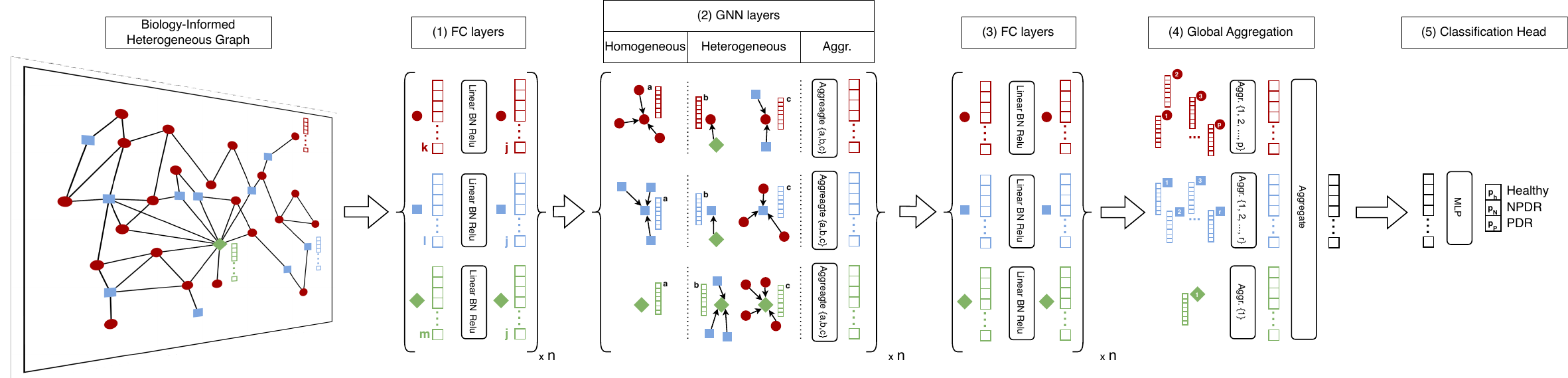}
    \caption{Illustration of the graph neural network architecture designed for the DR disease staging task. The architecture takes a heterogeneous graph as input and predicts the disease stage for the provided input graph. We generalize homogeneous message-passing architectures to the heterogeneous case by learning individual message-passing functions for each node type. After the message passing layers and linear layers, an aggregation for each node type is performed. Finally, the embeddings for all node types are aggregated and passed on to a classification head.}
        \vspace{-0.3cm}
    \label{fig:arch}
\end{figure*}

We utilize our heterogeneous graph representation to predict the DR stage of the imaged eye, which we frame as a graph classification task. We use the Pytorch Geometric framework \cite{fey2019fast} to configure established GNN layers for our presented heterogeneous graph structure. Specifically, we use the architecture depicted in Fig.~\ref{fig:arch}, which is composed as follows:

\begin{enumerate}
    \item First, we preprocess the different node types separately using a block of fully connected layers for each type. Each layer is followed by a batch normalization layer for regularization and a ReLU activation function, as shown in Fig. \ref{fig:arch} (1). These preprocessing layers can adjust the node embeddings for the following message-passing layer. 
    \item In the next step, we refine the embeddings based on neighborhood information using a block of independent message passing layers \cite{hamilton2017inductive} for each edge type, see Fig \ref{fig:arch} (2). The update using one specific type of neighborhood node, e.g., type $a$ is then defined as
    \begin{gather}
          h_{\mathcal{N}_a(v)}^k\leftarrow\mathrm{AGG.}_k(\{h_u^{k-1},\forall u\in\mathcal{N}_a(v)\}) \\
          h^k_{v_a}\leftarrow\sigma\left(W^k\cdot(h_v^{k-1},h_{\mathcal{N}_a(v)}^k)\right) .
    \end{gather}
    All edges are bidirectional and allow message passing in both directions. This message-passing layer is used to refine the node embeddings based on the neighborhood information stored in the graph. Using heterogeneous edges, node embeddings can be refined using the information of all types of connected neighbors. The resulting embeddings from the refinement using distinct types of neighborhood nodes are then aggregated 
    \begin{equation}
          h^k_v\leftarrow\mathrm{AGG.}_k(\{h^k_{v_a}, h^k_{v_b}, ..., h^k_{v_z}).
    \end{equation}
    \item After the graph layers, another block of fully connected layers postprocesses the individual representations of the different node types, see Fig \ref{fig:arch} (3). Again, batch norm layers and a ReLU activation function are applied following each layer. 
    \item Next, global pooling is applied to aggregate over the embeddings of all nodes of the same type. All the aggregated embeddings of the different node types, see Fig \ref{fig:arch} are then again aggregated into a single feature vector. (4). 
    \item Finally, we use an MLP head consisting of fully connected layers with dropout on the aggregated fixed-size feature vector for multiclass classification, see Fig \ref{fig:arch} (5). 
\end{enumerate}

Our proposed network is restricted in its complexity compared to current CNNs and transformer models. Table \ref{table:model_params} shows a comparison of the number of parameters and the training time of the GNN model and other prominent network architectures. Notably, the graph construction step is a prerequisite for our approach. We observe that with non-runtime-optimized code, the construction takes multiple seconds. Therefore, the benefit of faster training during hyperparameter tuning outweighs the construction overhead. 

\input{num_params}

\subsection{Interpretability Framework for Retinal Graphs}
Our explainability framework consists of two components. First, an attribution method, and second, a localization method. 

\subsubsection{Attribution Method}
We utilize integrated gradients \cite{sundararajan2017axiomatic, mccloskey2019using} as attribution method to describe the impact of specific nodes and their characteristics on the algorithmic outcome, following the evaluations by Agarwal et al. \cite{agarwal2023evaluating}. Alternatively, the explanations can be easily adjusted to any desired attribution method. The integrated gradients approach follows a straight line path in the input feature space from a baseline sample to the target sample. In computer vision, this baseline sample is typically a zero-image. Equivalent to baseline images, graph-based approaches require a baseline vector for each node in the graph. The attribution score $a_{v_i}$ of a feature $i$ of a node $v$ given a baseline $x_{bl}$ is then defined as 
\begin{align}
a_{v_i}::=(x_{v_i}-x^{bl}_i)\times\int_{\alpha=0}^{1}{\frac{\partial F(x^{bl}+\alpha\times(x-x^{bl}))}{\partial x_{v_i}}}d\alpha .   
\end{align}
Instead of using a constant baseline sample, such as a zero-vector, we propose a dynamic baseline \cite{sturmfels2020visualizing} that considers the unique characteristics of retinal vasculature, generating more fine-grained explanations. In contrast to natural images, the retinal vasculature adheres to a fixed structure that can be represented in the explanation process. Previous studies indicate variations in capillary density with increasing distance from the FAZ and across different retinal sectors \cite{lavia2020retinal}. This implies that distinct characteristics should be considered normal for intercapillary areas and vessels in different areas of the retinal vasculature. Therefore, we select the k-nearest same-type neighbor nodes from the training set and use their feature-wise mean as the dynamic baseline vector $x^{bl}$ for each node in a graph. Formally, the baseline vector $x_v^{bl}$ for a node $v$ is defined as

\begin{align}
    x_{v}^{bl} = \text{{avg}}\left(\left\{ x_{u} \, \middle| \, u \in \mathcal{N}_l(v) \right\}\right).
\end{align}

Here $\mathcal{N}_k(v)$ is the $k$ neighborhood of the node $v$. We define the $k$ neighborhood as the $k$ nodes of \textit{the same type} from the training set with the closest spatial correspondence to node $v$ and $x_u$ is the feature vector of $v_u$ after Z-score normalization. Using the k-nearest-neighbors for creating the baseline vector accommodates local variations at various distances from the fovea. We show an analysis on the spatial variation of vessel segment and intercapillary area characteristics in \ref{app:spatial_var}. We employ search trees to efficiently identify relevant neighbors in the training data.

\subsubsection{Localization Method}
For clinical utility, it is critical that graph nodes that have been identified as important (e.g., individual vessel segments) can be visualized in the corresponding OCTA image. These visualizations can be used to verify whether the extracted retinal characteristics align with the underlying image or if they arise from representation errors, such as image artifacts or segmentation errors. We store the corresponding segmentation map for each node in our graph representation to achieve such a localization. 

\subsubsection{Explanation Generation}
To generate our explanations, we first identify critical nodes using our attribution method by summing over the integrated gradients for the feature vector. The identified critical nodes are then spatially highlighted using the described localization method. Lastly, we identify the decisive characteristics of each critical node using the feature-wise integrated gradients. Examples of the generated spatial and feature-wise explanations are described in Section \ref{sec:results} and visualized in Figure \ref{fig:spatial_and_feature}.

\subsection{Datasets}
\subsubsection{Proprietary Clinical Dataset}
Our OCTA dataset contains 1268 OCTA images of the DVC. For some patients, images of both the right and left eyes are contained in the dataset. Beyond that, no patients are included more than once. DR grades are assigned on a patient level with the stages of \textit{healthy} (847 samples), diabetes mellitus (\textit{DM}) without DR (133 samples), early non-proliferative DR (\textit{early NPDR}, 134 samples), \textit{late NPDR} (62 samples), and \textit{PDR} (92 samples). The disease grades were assigned according to the International Clinical Disease Severity Scale \cite{wilkinson2003proposed}, relying on color fundus photography. The OCTA images are 3 mm $\times$ 3 mm, with a resolution of 304$\times$304 pixels.

We perform a stratified data split into six parts for our experiments. As sometimes both eyes of a single patient are imaged, we strictly keep both samples in the same split to avoid information leakage. One of the splits is kept as a test set, while the other five splits are used for cross-validation. In each split, \textit{healthy} and \textit{DM} are pooled into a single \textit{healthy} class, and \textit{early NPDR} and \textit{late NPDR} are pooled into a single \textit{NPDR} class. As is typical in a real-world setting, our datasets are imbalanced towards \textit{healthy} subjects. For these reasons, we evaluate and select our models based on balanced agreement.

\subsubsection{Public Dataset}
We use the OCTA-500 \cite{li2020octa} dataset as an external validation dataset for our method. For that purpose, we use the 200 3 mm $\times$ 3 mm images with a resolution of 304$\times$304 pixels and remove samples with labels other than \textit{DR} or \textit{healthy}. Labels in the OCTA-500 dataset are assigned based on medical patient records. This results in a total of 189 images, of which we use the ILM-OPL projection maps. Notably, these projections contain SVC and DVC vessels, resulting in a slight domain shift compared to our proprietary dataset. Of the 189 images, 160 are healthy samples, and 29 are in the DR class. We apply each algorithmic class's best-performing checkpoints to classify the OCTA-500 samples. Predictions of \textit{NPDR} and \textit{PDR} are pooled into a single DR class for comparison with the ground truth labels.\\

\subsection{Baseline Methods}
We compare our approach to the most prominent and well-performing approaches for DR staging in the literature. These methods include ML using pre-established biomarkers and DL approaches such as CNNs directly operating on the information-rich raw OCTA images. Furthermore, we compare our method to an ophthalmology foundation model that has not previously been employed for this task on OCTA images. Finally, we establish traditional ML methods on graphs that employ global feature aggregation followed by classic ML classifiers without using graph convolutions.

\subsubsection{Manual Biomarker Extraction}
We extract traditional biomarkers from the image segmentations. We use RF and SVM as classifiers for the DR stage prediction due to their excellent performance in a wide range of applications for tabular data. Specifically, we extract the following biomarkers and use them as classification features: 
\begin{enumerate}
    \item FAZ biomarkers: area, maximal diameter, mean diameter, acircularity index, STD4, and NR300. The reader is referred to \cite{lu2018evaluation} for detailed descriptions of the last three biomarkers. 
    \item Vessel and intercapillary area biomarkers: vessel density, vessel perimeter, and fractal dimension as a vessel complexity measure. 
\end{enumerate}

\subsubsection{Vision Models}
We compare to well-established CNN architectures previously used for the task at hand \cite{qian2023drac}. We implemented EfficientNet \cite{tan2019efficientnet}, ResNet \cite{xie2017aggregated}, and VGG \cite{simonyan2014very} models in combination with an optimized data augmentation pipeline. We use all models with weights pre-trained on ImageNet. Beyond these networks, we use the more recent ConvNeXt \cite{liu2022convnet}, Efficient Net V2 \cite{tan2021efficientnetv2}, and SwinTransfomer \cite{liu2021swin} architectures. We use cross-entropy loss for CNNs and all other neural network architectures.

\subsubsection{RETFound Foundational Model}
The RETFound model \cite{zhou2023foundation} is built on a vision transformer architecture. To initialize this complex architecture's weights, the model is pre-trained in a self-supervised fashion on more than 900,000 color fundus photography images before being fine-tuned on our training set. We decided to use the color fundus pre-trained model instead of the OCT fine-tuned models due to better performance. This is likely caused by the larger domain shift from en-face OCTA to OCT than to color fundus photography.  

\subsubsection{Traditional ML on Node Embeddings}
Traditional ML on graphs methods are applied to assess the feature quality of node embeddings in our heterogeneous graph. Global aggregation generates a single feature vector for vessel and intercapillary area nodes and the FAZ node. These aggregates are then concatenated and used as input for ML classifiers, with RF and SVM chosen for their effectiveness. 

\subsection{Metrics}
We report \textit{ROC-AUC} for \textit{no DR}, \textit{NPDR}, and \textit{PDR} class staging and for the binary \textit{DR vs no DR} classification. The \textit{class balanced agreement} metric indicates how often the clinical fundus labeling and the OCTA-based stage prediction match. \textit{F1} scores are used to indicate the class-specific agreement rates. Perfect agreement rates for clinical fundus labels and 3mm x 3mm microvascular information are neither expected nor desired. This would disregard the additional microvascular information from OCTA imaging, see Figure \ref{fig:octa_v_fundus} and section 4.5.

\input{results1}

\section{DR Staging Experiments}
\label{sec:results}
\subsection{Agreement Rates of Machine Learning OCTA Grading Approaches}
Our biology-informed heterogeneous graph representation achieves comparable ROC AUC scores of 0.829, matching the best CNN (0.822) and transformer models (0.819). Staging purely based on established biomarkers reaches notably lower agreement with the CFP-based labels. We attribute the lower agreement to the lack of biomarkers that fully encode pathological information on intraretinal microvascular abnormalities and microaneurysms. In contrast, CNNs and our heterogeneous graph representation plateau at agreement rates where the DR stage is determined by microvascular information in OCTA images that diverges from clinical evaluation through CFP (see Figure \ref{fig:octa_v_fundus}).

\input{results2}

\subsection{Sensitivity analysis on OCTA image variations and segmentation maps}
A comparison of using the SVC instead of the DVC using the same CNN architecture shows similar staging performance for using either the DVC or the SVC layer. Moreover, we experimented with directly using segmentation maps as input to the CNN classifier. Interestingly, this results in comparable performance to raw images and opens an interesting avenue for enhancing explainability in classification with CNN models. 

\subsection{Generalization to public OCTA-500 dataset}
We use the checkpoints with the highest class-balanced agreement rates from the 5-fold cross-validation on the proprietary dataset for evaluation on the OCTA-500 dataset. Our proposed graph learning method yields the highest agreement with the medical record labels in this 0-shot evaluation on the OCTA-500 dataset (see Table \ref{table:octa500}). This highlights the robustness of our approach to minor domain changes induced by different hardware (OCTA scanners) and preprocessing algorithms (e.g., layer segmentation and intensity clipping). The performance between the CNN models and the RETFound model differs strongly, probably due to differences in the high-level concepts used by the vision models. Due to their lack of interpretability, the underlying mechanisms that cause the model to fail cannot be identified. The simple random forest models perform excellently in this generalization task. VGG11 and ResNet18 exhibit high AUC ROC values compared to their relatively low balanced agreement. This indicates that while these models are successful in differentiating the disease level, the classification threshold does not generalize to the new dataset.

\input{results3}

\subsection{Comparison to Homogeneous Graph Representations}
We experimented with a simplified setting of the individual homogeneous graph representations and included the results below in Table \ref{table:homog}. The intercapillary region graph does not achieve comparable agreement rates measured by ROC AUC (0.815), compared to both vessel (0.828) and heterogeneous graphs (0.829). We attribute this to the lack of information about intraretinal microvascular abnormalities and microaneurysms. The performance of the vessel graph is comparable to the performance of the heterogeneous graph. In fact, the vessel graph can indirectly encode the crucial pathological non-perfusion information through a reduction in total vessel area. However, only the heterogeneous graph allows for a direct attribution to the non-perfusion regions and therefore the absence of capillaries.

\begin{figure}
    \centering
    \includegraphics[width=0.8\linewidth]{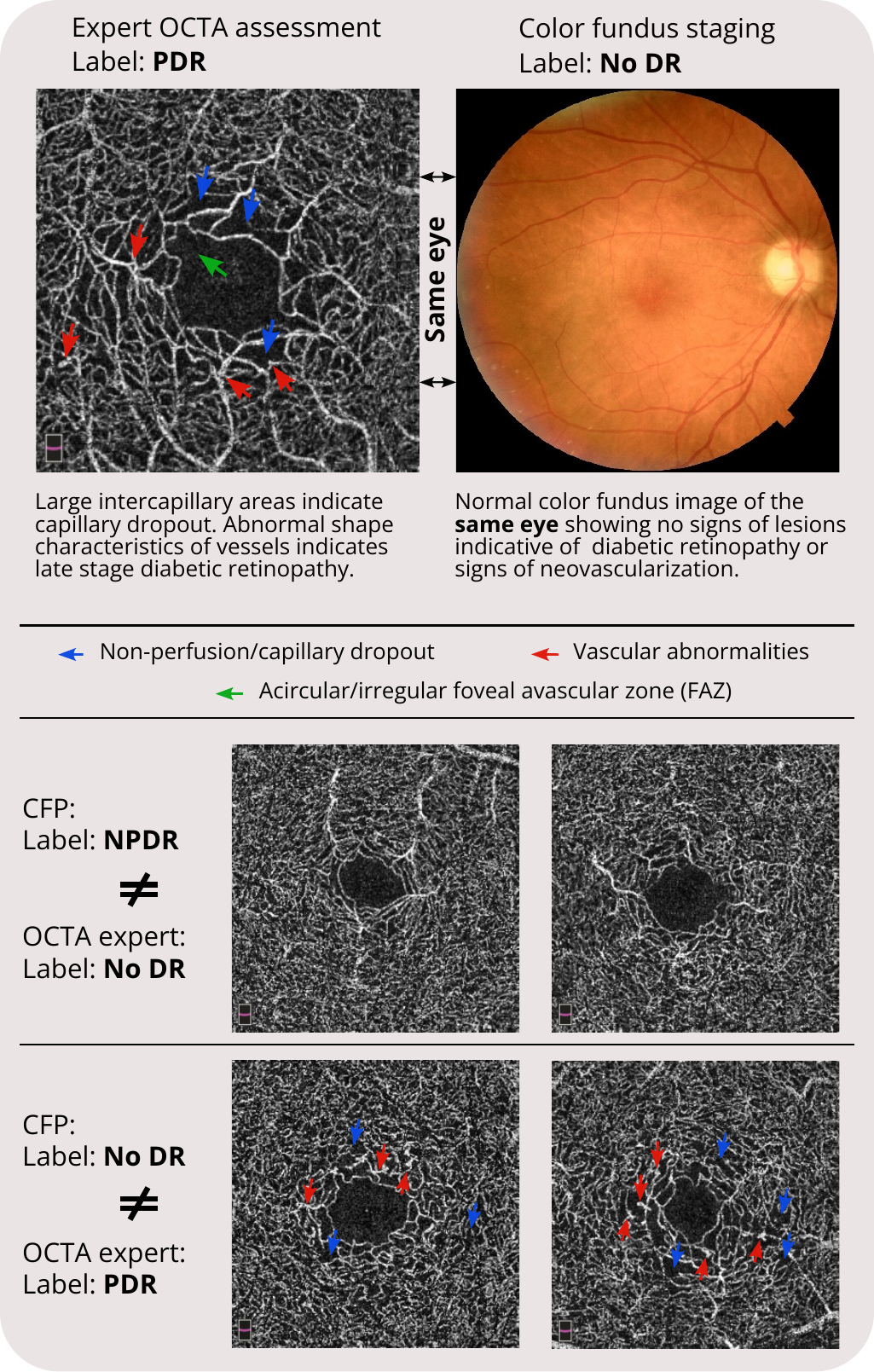}
    \caption{Overview of cases where the microvascular changes assessed using en face OCTA images do not align with the disease progression information obtained from color fundus photography (CFP). (top row) Comparison of an en face OCTA image with the matching CFP image (middle and bottom row). Further cases of OCTA images where the OCTA-based expert assessment does not align with CFP grading.}
    \label{fig:octa_v_fundus}
\end{figure}

\subsection{Expert reading of OCTA predictions that disagree with clinical fundus stage labels} 

DR stage labels in the proprietary dataset were assigned according to the International Clinical Disease Severity Scale \cite{wilkinson2003proposed} based on the Early Treatment Diabetic Retinopathy Study (ETDRS), which relies on color fundus photography. This feature-based grading approach does not necessarily align with the microvasculature changes visible in OCTA images \cite{thompson2019optical}. Pathological microvascular alterations can occur without microaneurysms observable on fundus photography, a phenomenon we encountered frequently in our dataset. 

To investigate these staging discrepancies, we performed staging experiments with a senior ophthalmologist with over ten years of OCTA and DR experience. Nine cases were preselected where OCTA model predictions plausibly disagreed with fundus labels (see Section 5). Expert review confirmed the disagreement in eight of these cases. We report cases where expert ophthalmologists assess OCTA images as NPDR or even PDR stage, while no retinopathy is detected in the corresponding fundus images (see Figure \ref{fig:octa_v_fundus} first and last row). Conversely, sometimes no microvascular abnormalities are visible in the spatially limited 3 mm x 3 mm OCTA image, while the color fundus examination indicates the DR (see Figure \ref{fig:octa_v_fundus} middle row). 

In clinical practice, such modality discordance between fundus and OCTA labels is particularly relevant, as clinicians must reconcile information from both imaging techniques to formulate unified staging and treatment decisions.

Despite expert assessments of both Fundus and OCTA images that highlighted staging discrepancies, the study has limitations, including the number of cases assessed and the lack of multi-site evaluation. These limitations warrant further investigation in a large scale study with multi-site evaluation.
\section{Interpretability Analysis and Explainability of Staging}
\label{sec:explanations}

\subsection{Feature Relation to Microvascular Pathologies}
\begin{figure*}
    \centering
    \includegraphics[width = \textwidth]{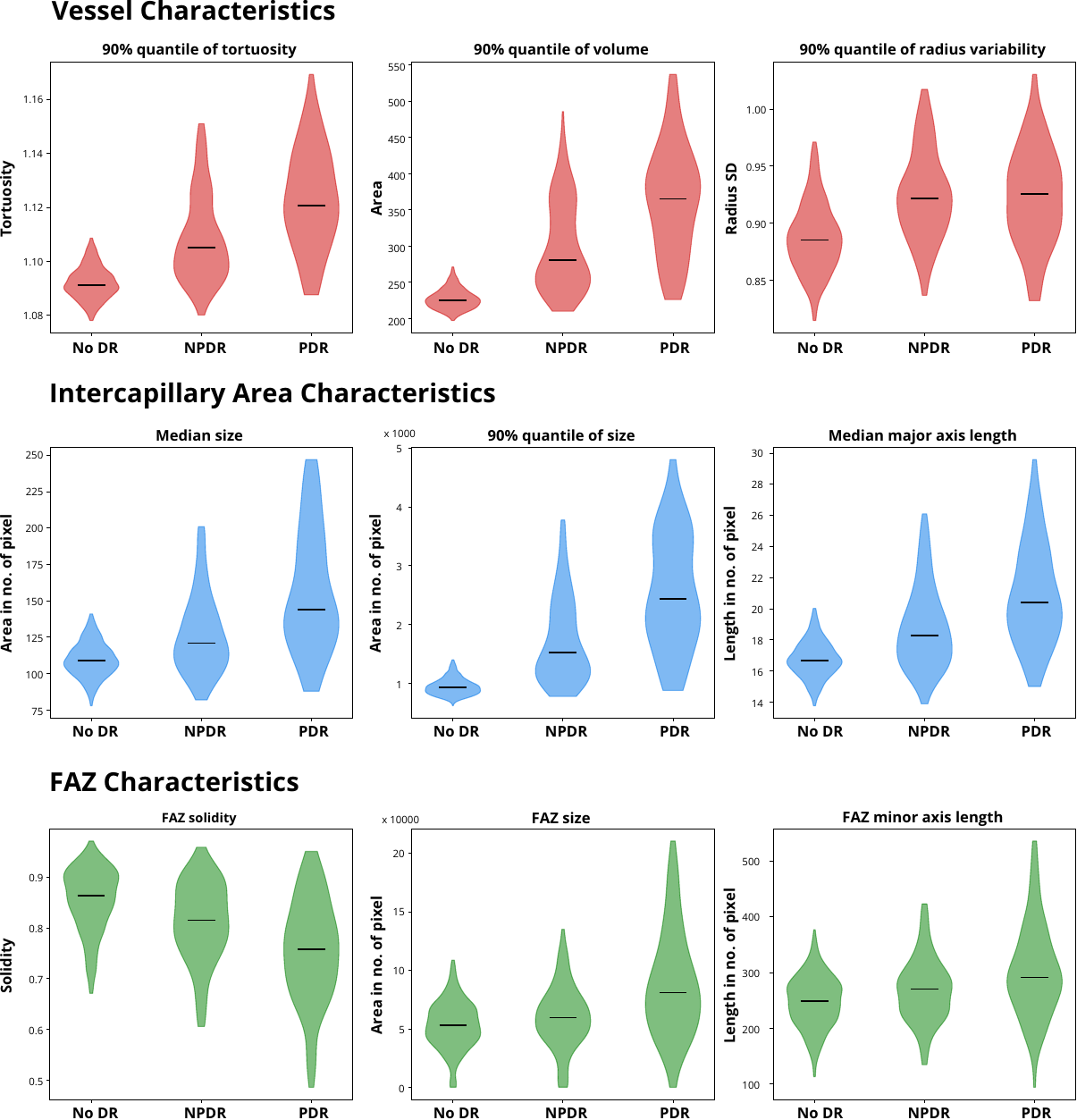}
    \caption{Analysis of specific features that are encoded in our heterogeneous graph representation for vessel segments (top row), intercapillary areas (middle row), and the FAZ (bottom row). Violin plots are stratified by clinical DR grade. The violin plots show the distribution of a specific characteristic (e.g., median intercapillary area size) across all OCTA scans of the specified disease stage.}
    \label{fig:features}
\end{figure*}

Our feature embeddings for intercapillary areas, vessel segments, and the FAZ are directly linked to pathological changes of the microvasculature. Figure \ref{fig:features} illustrates the relationship between specific features and the progression of DR. The increase in tortuosity of vessels is defined as a microvascular abnormality; we observe a clear increase in the 90\% quantile tortuosity of vessel segments (Figure \ref{fig:features} top left). Capillary degeneration results in less branching and therefore larger vessel segments; moreover, microaneurysms and neovascularizations often appear as large segmented vessel segments. An increased 90\% quantile for vessel segment volume clearly reflects these pathological changes (Figure \ref{fig:features} top center). Microvascular abnormalities and microaneurysms are characterized by irregular shapes. We observe that the 90\% quantile of the radius variability of isolated vessel segments is increased for the NPDR and PDR stage (Figure \ref{fig:features} top right). Capillary loss is best reflected in the characteristics of intercapillary areas. We see a clear increase in the median area of intercapillary areas (Figure \ref{fig:features} middle left), and even more drastically for the 90\% quantile (Figure \ref{fig:features} middle center). This capillary loss is also reflected in other characteristics, such as the increased median major axis length of intercapillary areas (Figure \ref{fig:features} middle right). Changes to the FAZ characteristics, due to the loss of integrity of parafoveal vessels, are a known indicator for the progression of DR. We see this effect most pronounced in the decreased solidity (defined as the ratio of the area of a region to the area of its convex hull) (Figure \ref{fig:features} bottom left). Changes, although less pronounced, are also visible in the area and the length of the minor axis of the FAZ (Figure \ref{fig:features} bottom center and right).

\begin{figure*}[!b]
   \centering
    \includegraphics[trim={0cm 0cm 0cm 0cm},clip,width = \textwidth]{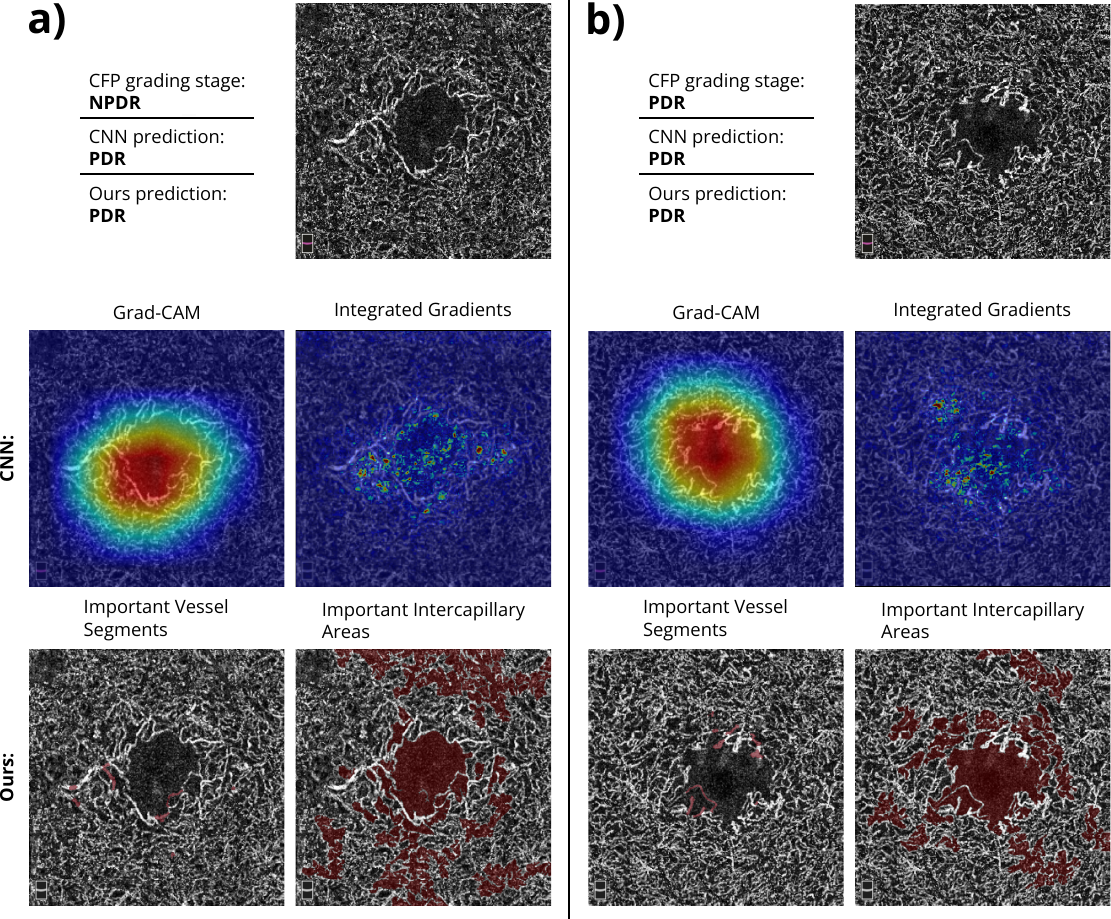}
    \caption{Comparison of different explainability tools for the DR staging task on OCTA images. GradCAM and integrated gradients are state-of-the-art approaches that previous works have applied to DR staging. 
    Sample \textbf{a)} describes a sample with clinical diagnosis \textit{NPDR} that the CNN and graph classification model both classify as \textit{PDR}. Sample \textbf{b)} describes a sample with clinical diagnosis \textit{PDR} that the CNN and graph classification model both also identify as \textit{PDR}. Compared to traditional spatial explanations for vision models, our framework (bottom row) enables the precise localization of critical vessel segments with shape and intensity that indicate intraretinal microvascular abnormalities and perfused microaneurysms. Intercapillary areas indicate image areas where our model sees reduced capillary density. }
    \label{fig:spatial_explanations}
\end{figure*}

\subsection{Examples for spatial explanations}
To illustrate that (1) our method links pathological microvascular changes to the stage prediction task, and (2) our method allows traceable prediction results, we show explanations for predicted DR stages. Our method allows the spatially exact identification of the most important vessel segments and intercapillary areas for the DR stage prediction task. Figure \ref{fig:spatial_explanations} shows two examples where the most important vessel segments and intercapillary areas are highlighted for two patients suffering from DR. The precise spatial attribution is in stark contrast to common CNN explanations, which often provide broad and scattered explanations that do not allow for a precise attribution to abnormal vessels, areas of capillary dropout, or microaneurysms, guiding the prediction. We adopt explainability approaches previously employed by Ryu et al. \cite{ryu2021deep} and Heisler et al. \cite{heisler2020ensemble} and use grad-CAM to create model explanations. The grad-CAM explanations are focused on the center of the FAZ, which is unrelated to DR disease progression. Similarly, a majority of the area highlighted by integrated gradients, combined with the vision models, highlights areas within the FAZ. For the subject in \ref{fig:spatial_explanations} a) where the model prediction on OCTA deviates from the clinical fundus grading, both vision explanations do not provide an interpretable explanation for the prediction of the further progressed DR stage. In contrast, our methods clearly highlight vessel segments that resemble abnormal structures that guide the model to predict late-stage DR, deviating from the fundus-based diagnosis.

\begin{figure}
    \centering
    \includegraphics[width=0.85\linewidth]{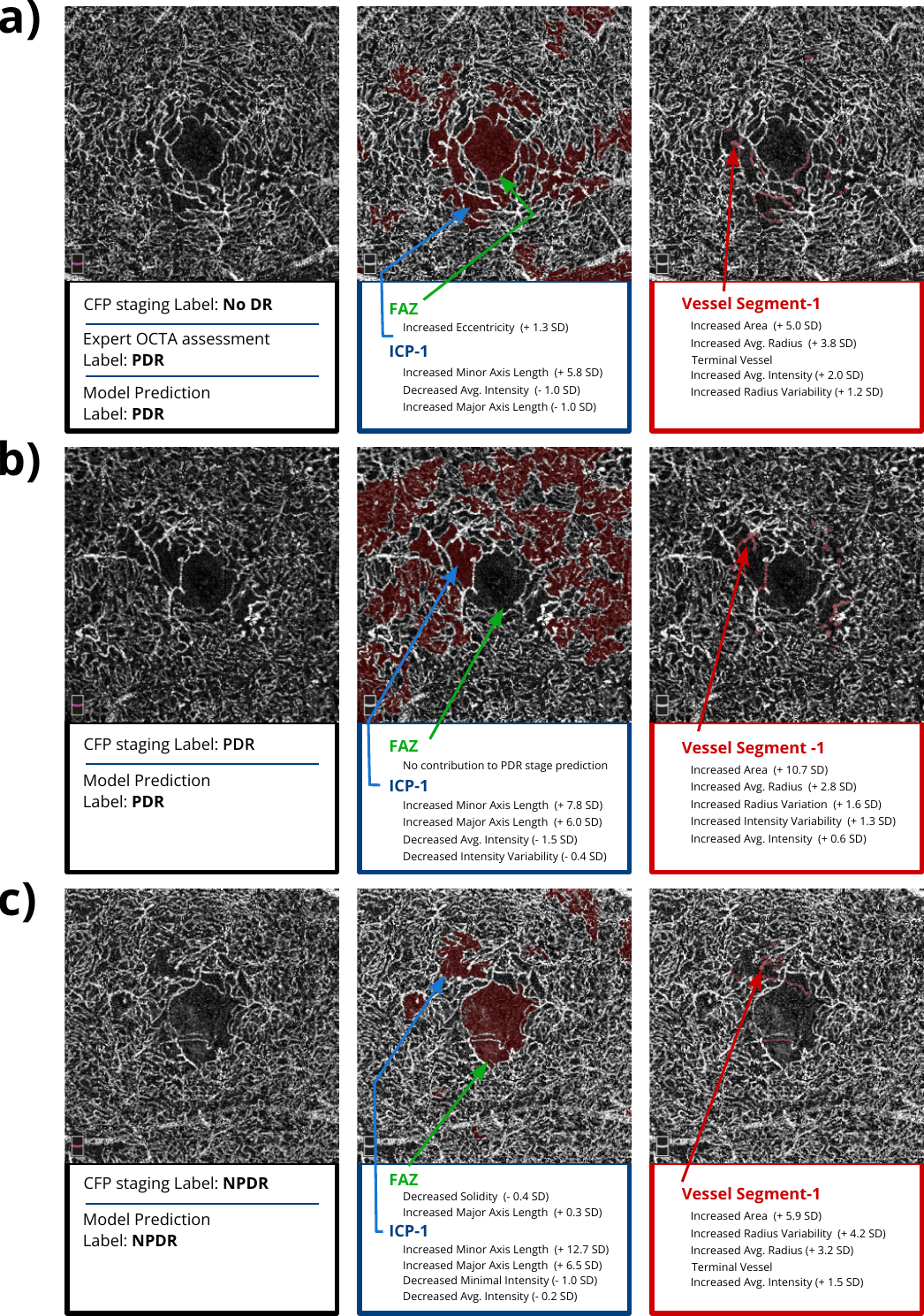}
    \caption{Explanations for model predictions and explanations of our framework with predicted stages agreeing to color fundus staging (b), c)), and an example with diverging prediction (a)). Red overlay indicates contribution of intercapillary areas, FAZ (center column), or vessel segments (right column) to the predicted stage. Boxes below the images indicate which features are most important to the prediction, and how they diverge from the norm.}
    \label{fig:spatial_and_feature}
\end{figure}

Beyond the spatial explanations, our method enables us to attribute the elevated importance of specific nodes to high-level concepts, such as vessel tortuosity or intercapillary area size, that are directly linked to disease progression. We display three different examples: Figure \ref{fig:spatial_and_feature} a) shows an example where the clinical label is no DR while the OCTA-based prediction as well as the OCTA expert label indicate PDR, Figure \ref{fig:spatial_and_feature} b) displays an image that was classified as PDR with clinical label PDR, and Figure \ref{fig:spatial_and_feature} c) displays an image that was classified as NPDR with clinical label NPDR. We indicate the most important characteristics and their deviation from the norm for exemplary nodes. Additionally, we provide full interactive explanations in our code repository, which allows us to see the influence of intercapillary areas, the FAZ, vessel segments, and their characteristics on the predicted DR stage. 

Example a) in Figure \ref{fig:spatial_and_feature} shows the OCTA image of a subject with no DR following standard fundus examination. The OCTA image was graded with the PDR stage by an expert ophthalmologist. Our model aligns with the OCTA expert analysis, highlighting the intercapillary areas around the FAZ as highly important, due to their increased size (e.g., minor axis length and major axis length) and their low average image intensity. The eccentricity of the FAZ also contributed to the stage prediction. Moreover, specific vessel segments contribute to the PDR diagnosis, such as the bright intensity terminal vessel emphasized by the red arrow. In the prediction, the size (area and average radius), shape abnormality (increased radius variability), hyperintensity, and the terminal position are the most important characteristics.

Example b) in Figure \ref{fig:spatial_and_feature} shows an example where the clinical stage is PDR, aligning with the severity of the microvascular changes in the OCTA image. While the FAZ does not indicate the PDR stage, the large size and low intensity in the capillary areas are indicative of the PDR stage. Moreover, the model identifies multiple abnormally shaped vessels, such as the vessel segment indicated by the red arrow, which exhibits increased size (area + radius), atypical shape characteristics (increased radius variability), and unusually high variation of intensity within the segment. 

Finally, example c) in Figure \ref{fig:spatial_and_feature} shows the OCTA image of a subject with clinical label NPDR and model prediction NPDR. The decreased solidity of the FAZ and the slightly increased (+0.3 SD) major axis length contribute to the NPDR prediction. Moreover, some intercapillary areas contribute to the DR prediction through increased size and decreased intensity. The vessel segment indicated by the red arrow, which shows strong similarities to a microaneurysm, strongly contributes to the NPDR diagnosis through size (area and radius), shape (radius variability), hyperintensity, and terminal position. 

Notably, the baselines, as depicted in Figure \ref{fig:spatial_explanations}, are limited to spatial explanations. In contrast to our framework, they do not offer feature-wise attributions towards interpretable concepts (see Figure \ref{fig:spatial_and_feature}). Therefore, our comparison is limited to these spatial explanations. 

\subsection{Agreement of explanations with expert annotations}
To quantitatively assess the interpretability advantages of the biology-informed graph representation for disease staging, we compare spatial expert annotations for abnormal vessel segments and non-perfusion regions against attribution maps from baselines and the graph-based staging. The expert marked the structures that are representative of the disease stage: in NPDR eyes, vessel abnormalities and non-perfusion regions that indicate the presence of DR; in PDR eyes, only those whose severity or extent indicate PDR rather than NPDR. For all models, attribution maps are computed with respect to the OCTA-assessed disease stage, so that model evidence and expert evidence are compared for the same class. Attributions of our graph nodes are mapped to pixels using the localization method described in Section 3.3.2. We calculate pixel-level AUPRC and relevance mass accuracy (RMA) \cite{arras2022clevr} for abnormal segments and non-perfusion areas to measure how well the attribution concentrates on the annotated structures. A detailed description of both metrics is provided in \ref{app:expert_annotations_metrics}. The chance level reported in Table \ref{tab:interpretability_combined} is the AUPRC of a uniformly random attribution map, which equals the fraction of annotated pixels in the image: 0.007 for abnormal vessels and 0.030 for capillary dropout.

The results of the analysis are reported in Table \ref{tab:interpretability_combined}. Comparisons are performed against six CNN models with both GradCAM and IG attribution maps. The graph-based approach achieves the highest AUPRC and RMA values across all comparisons. All methods remain clearly below perfect attribution, especially for abnormal vasculature, whereas scores are better for dropout areas. Introducing a vessel-segmentation-based attribution restriction for the vision model attributions improves AUPRC and RMA scores, especially for the attribution quality of abnormal vasculature. This is similar to what the graph-based approach is inherently doing through its data structure. The results of this analysis are displayed in Table \ref{tab:interpretability_combined_restricted}. We discuss the limitations of this analysis in Section 7.

\input{r2_quantitative_interpretability_combined}

\subsection{Comparison to Other Graph-Based Explanations}
Random forest and support vector machine classifiers on aggregated node embeddings come with the disadvantage of losing spatial information for explainability. The aggregation of node information followed by an ensemble of decision trees in random forests does not allow spatial hinting towards critical regions. 

Homogeneous graph representations do not allow for a direct attribution towards clinically well-studied microvascular changes. Using vessel graphs alone does not allow direct attribution to regions of ischemia and capillary dropout, which are hallmarks of DR. Using intercapillary areas alone does not allow attribution to intraretinal microvascular abnormalities, which are the second pillar for the analysis of DR in OCTA images. Only the combination of both representations provides the capability to attribute the classification directly to the microvascular changes that comprise dropout, abnormalities, and new-growth of vessels. 

\section{Stability Analysis}
\label{sec:stability}
Our approach relies on segmentation and graph extraction before graph representation learning for disease staging. While this approach sets it apart from standard vision model processing in terms of better interpretability, it introduces failure points at intermediate processing steps. We perform a detailed step-wise analysis of the expected variance at all processing steps for a realistic test-retest scenario.

For this purpose, we rely on proprietary repeat scans of 30 eyes from 15 healthy participants, acquired in a real test-retest scenario, with 5 repeated measurements (150 images total). The demographics of this cohort and, for comparison, of the main clinical cohort are reported in Appendix A.4, Table \ref{tab:cohort_demographics}. The distortion types and parameter ranges of our synthetic scenario (Table \ref{tab:distortion_params}) were calibrated through visual comparison against the variations observed between these repeat acquisitions (see Appendix A.3). We then augment our existing data to simulate the test-retest scenario and run the full analysis pipeline on both the original and augmented samples. The resulting stability estimates are best interpreted as an upper bound. The limitations of the synthetic test-retest scenario are discussed in Section 7.

\subsection{Segmentation Stability}
We segment the original and five augmented versions of each image in the dataset. After segmentation, we assess the reproducibility of the segmentation map using the Dice coefficient. We reverse the geometric transformations to ensure registered maps before calculating the Dice score. To account for the boundary deviations, we compare center crops of 2.4 mm × 2.4 mm. The results are displayed in Table \ref{tab:dice_foreground_combined} in addition to the foreground ratios stratified by disease stage. The average Dice score for the vessels in the test-retest scenario is ~97\%, indicating good reliability of the segmentation maps. Similarly, the overlap of the intercapillary areas, defined by the inverse map, is high (97\%). Dice scores for vessel segments become slightly lower with increasing disease severity (No DR: 96.8\%, PDR: 96.1\%), while scores for the intercapillary areas remain stable across stages.

\input{stability_segmentation}

\subsection{Graph Similarity}
In the next step, we compare the similarity of graph representations extracted from the test-retest segmentation maps. We quantitatively estimate the similarity of graph representations using counts of vessel and intercapillary area nodes (Table \ref{tab:combined_graph_ica_2D_TTA}). Vessel segment counts show an average test-retest standard deviation of 2.1\%, compared to an inter-patient standard deviation of 14.6\%. The test-retest variability is thus substantially lower than the inter-patient variability — the graph representation captures disease-relevant differences between patients while remaining stable under scan-to-scan variations within patients. As for the segmentations, variability increases with disease severity: test-retest SD rises from 1.8\% for No DR to 4.2\% for PDR for vessel segments. This pattern is expected, as diseased vasculature with capillary dropout and irregular structures is more sensitive to imaging variations. Intercapillary areas show a similar pattern.

\input{stability_graphs_v2}

\subsection{Classification and Explanation Stability}

To assess the impact of input data variance on disease staging, we classify the augmented test data using our models. For this purpose, we train a model on the same splits as previously used, extended with test-retest data. We quantify the stability of stage-label predictions using two metrics: the agreement rate, which measures the fraction of predictions matching the majority vote across all augmented versions; and the consistency rate, which measures the fraction of augmented predictions matching the original sample's prediction. Results are shown in Table \ref{tab:stability_validation}.

Overall consistency is high (90.6\%), with No DR and PDR showing the highest stability (91.6\% and 90.6\%, respectively). NPDR shows lower consistency (82.9\%), which is clinically plausible: as the intermediate stage, NPDR cases often lie near decision boundaries and may shift to adjacent stages under scan variations. To characterize the severity of prediction deviations, Table \ref{tab:stage_deviation_validation} shows the distribution of stage deviations. The vast majority of deviations are single-stage (9.0\%), while two-stage deviations — where a scan shifts from No DR to PDR or vice versa — are rare (0.4\%). This indicates that even when predictions change, they typically remain clinically proximate.

\input{stability_classification}
\input{stage_deviations}

We generated explanations highlighting the most important vessel segments and intercapillary areas for the synthetic test-retest data to investigate the stability of explanations. Figure \ref{fig:explanation_stability} shows the explanations for three samples with the original image and three synthetic retest images. The experiment shows that important vessel branches are mostly consistent. However, the exact spatial outline of the vessel segments often differs between the synthetic scan variations. We attribute this to variations in the vessel segment compartmentalization from different segmentation maps during the vessel graph generation. Moreover, some segments/areas are highlighted in some scans and missing in others; this is linked to the importance visualization threshold used for visualization. To keep the explanations uncluttered, we only highlight segments and areas above a fixed importance threshold, which results in some areas/segments sometimes being absent when they fall below the threshold.

\section{Discussion}
\label{sec:discussion}
In this work, we propose an interpretable method for DR staging support using OCTA images. We establish a specialized heterogeneous graph representation that models the particularities of the retinal vasculature. Building on this novel data representation, black{we configure an established, efficient GNN model} that solves DR staging as a graph classification task. Finally, we introduce an interpretability tool providing fine-grained explanations for our predictions using spatial explanations that highlight critical structures and attribute their importance to specific characteristics. 

Our staging approach provides a notable step towards interpretable disease staging of OCTA images, which is crucial for adapting discriminative models in clinical practice. 
While our method improves existing explainability frameworks, we have not created a perfectly interpretable classification pipeline. Our improvements concern perspective and mathematical interpretability. Perspective interpretability is improved by the fine-grained spatial explanations that allow us to precisely visualize important vessels and intercapillary areas, which has been impossible for CNNs and vision transformers. Moreover, our attributions consist of human-interpretable concepts that are commonly associated with disease progression. By introducing a data representation based on the retinal vasculature, we restrict our model to these anatomical constraints in decision-making. We improve mathematical interpretability by limiting learnable parameters and network layers \cite{barcelo2020model}. Nevertheless, our model consists of multiple layers with activations and numerous features, which allows for non-trivial feature interaction. Therefore, the study of feature interaction, which has been previously done for MLPs \cite{tsang2017detecting}, is a necessary future work.

Although our approach avoids using the largely uninterpretable CNN and transformer models for the classification stage, it still relies on these architectures to generate high-quality segmentations. Utilizing this class of networks in the data representation stage provides some advantages compared to the usage at a later stage. The first advantage is the enhanced interpretability of the algorithmic outcomes using the abstracted representation. The second advantage is that the vision models can be trained on existing large-scale synthetic datasets \cite{kreitner2023detailed} compared to the disease classification task, where high-quality data is still limited. 
A limitation persists in the susceptibility of the graph representation to segmentation errors. To quantify the effects of these segmentation errors, we performed extensive experiments with distorted raw images and quantified the effect on the segmentation mask and the rest of the classification pipeline. While the variability is relatively low, there are still samples in which these changes are not negligible. 
Segmentation errors affect not only the predictions but also the validity of the resulting explanations, particularly in artifact-prone cases. For example, projection artifacts can produce segmented vessels that end abruptly and, viewed in the segmentation alone, resemble an intraretinal microvascular abnormality or a microaneurysm. An explanation attributing the prediction to such a structure remains faithful, since it correctly reflects what drove the model's decision, but is no longer accurate, as the highlighted structure is not a true indicator of DR but rather a consequence of the erroneous underlying representation. Our localization method (Section 3.3.2) provides a practical safeguard for exactly these cases: because every attributed graph node can be projected back onto the raw OCTA image, clinicians can visually verify whether a highlighted structure corresponds to true anatomy or to a segmentation or imaging artifact, a verification that is not possible for the internal representations of existing end-to-end vision models. This is a problem that warrants further work on creating topologically robust segmentations and graph representations. We argue that single-stage CNN or vision transformer models would implicitly create internal representations similar to a segmentation and, therefore, may be similarly affected. Moreover, segmentation errors are easier to detect than reasoning about erroneous internal representations in CNNs and vision transformers.

Robustness and interpretability are distinct requirements: Section 5.3 evaluates how well our explanations agree with expert annotations compared to vision-based methods, whereas Section 6 evaluates the stability of predictions and explanations under scan-to-scan variation.

The quantitative comparison of explanations in Section 5.3 is also subject to limitations. It rests on annotations from 11 eyes by a single expert, leaving the estimates sensitive to the reader and the limited samples. The inter-rater agreement on such annotations is unknown. A larger, multi-rater annotation study would allow the margins observed here to be tested statistically.

Several limitations apply to the stability analysis presented in Section 6. First, the repeat scans were acquired only from healthy and substantially younger participants. Test-retest variability in eyes with diabetic microvascular pathology may be larger, so stability estimates for diseased eyes should be interpreted with additional caution. Second, the repeat scans stem from a single device (Heidelberg Spectralis) operated by a single operator, with repeat acquisitions performed within the same session. Different devices, operators, and scanning days introduce additional sources of variability, so real-world test-retest fluctuations are expected to differ from those characterized here. Third, the distortion parameters were tuned through visual comparison against this small repeat-scan dataset (Appendix A.3). The ranges in Table \ref{tab:distortion_params} are therefore a visually validated approximation of test-retest variability. Fourth, the synthetic scenario does not reproduce all sources of real-world variability. Severely artifact-affected scans were excluded from the main dataset, and none occurred in the test-retest cohort, so the variability introduced by such acquisitions is not reflected in either dataset. Moreover, the augmentation pipeline cannot mimic all OCTA-specific artifact types, e.g., varying degrees of projection artifacts from vessels in more superficial complexes, motion artifacts, or retinal layer-segmentation errors. Consequently, the reported stability estimates reflect performance under favorable imaging conditions and constitute an upper bound. Robustness may be lower for artifact-affected scans, other acquisition settings, and diseased eyes. An empirical test-retest study involving patients with DR and multiple devices and operators remains an important direction for future work.

Our dataset covers a relatively homogeneous subgroup in which inter-patient variability is limited. Previous studies show that retinal characteristics detectable by OCTA vary significantly across ethnic groups 
\cite{laotaweerungsawat2021racial}. These subgroup differences need to be 
taken into account for application to new subgroups. Subgroup-specific 
differences, such as different average FAZ areas 
\cite{laotaweerungsawat2021racial} and within-group standard deviations, 
could in principle be addressed through subgroup-specific feature 
normalization, although this approach remains to be validated in practice.

Beyond image classification, we believe that our representation provides a foundation for future applications, such as link prediction to correct erroneous segmentation and node classification of intercapillary areas or vessel segments, e.g., for the identification of microaneurysms, intraretinal microvascular abnormalities, and neovascularization.

As discussed in the introduction, OCTA-derived microvascular 
information has demonstrated prognostic value for DR progression in multiple 
prospective studies. We envision our interpretable staging approach as a 
component within this complementary diagnostic workflow, where traceable 
microvascular explanations can support clinicians in identifying patients 
who warrant closer follow-up.

\appendix
\section{Acknowledgements}
\subsection{Funding}
This research did not receive any specific grant from funding agencies in the public, commercial, or not-for-profit sectors.
\subsection{Generative AI}
We used large language models at the sentence level to improve formulations and correct spelling and grammar. 

\subsection{Test-retest Scenario Augmentations}

We show the augmentation parameters for our synthetic test-retest robustness analysis in Table \ref{tab:distortion_params}. Each distortion type mimics one source of intra-patient scan-to-scan variability observed between the empirical repeat scans. Rigid transformation models residual misalignment between repeat acquisitions; translation is reported in pixels of the $1216 \times 1216$ image (upscaled raw images used as segmentation input) and rotation in degrees. The spatial gradient models signal strength differences across the field of view. The strength denotes the relative change in intensity from the image center to the edge along a linear ramp, and the angle denotes the orientation of that change. Gaussian noise models sensor- and speckle-related noise differences. The standard deviation is reported in 8-bit intensity units. Brightness/contrast models global intensity shifts (additive offset) and contrast scaling (multiplicative factor) between acquisitions. Signal strength models global differences in OCTA signal, applied as $I \mapsto 255\,(I/255)^{1/s}$ with $s$ the reported gamma factor. Signal attenuation models localized loss of signal: in 20\% of augmented samples, a rectangular region of 40--119 pixels side length is scaled by 0.3--0.7. The parameter ranges were determined through visual calibration against empirical repeat scans. We applied candidate augmentations to the test-retest samples and tuned the distortion types and parameter ranges until the synthetic deviations matched the deviations between real repeat acquisitions in appearance. Five distortion instances were drawn uniformly at random from these ranges, one value per distortion type, and applied jointly to every scan, yielding five matched counterparts per scan. Results of the experiments with this test-retest data are depicted in Section 6.

\input{stability_distortion_parameters}

\subsection{Cohort Demographics and Acquisition Protocol}
\label{app:cohort_demographics}

\subsubsection{Cohort Demographics}
Table~\ref{tab:cohort_demographics} reports the demographic characteristics of the test-retest cohort used to calibrate the synthetic scenario, alongside those of the main clinical cohort, stratified into no DR and DR classes. The test-retest cohort is substantially younger than the main cohort and, by design, free of DR. The implications for the representativeness of the calibration data are discussed in Section~\ref{sec:discussion}.
\input{r2_cohort_demographics}

\subsubsection{Acquisition Protocol}
The test-retest cohort comprises only healthy adults without any relevant ocular pathology and clear optic media. The subjects were measured five times within a single session using a Spectralis device (Heidelberg Engineering, Heidelberg, Germany). Scans with strong movement or shadow artefacts would have been excluded from the test-retest cohort. No scan required exclusion, so all 150 images were retained. The main dataset was acquired using an RTVue-XR Avanti device (Optovue Inc., Fremont, California, USA) from healthy patients, patients with diabetes mellitus without clinically diagnosed DR, and patients with varying degrees of DR. No patients with other retinal or choroidal diseases were included in the dataset. Eyes with significant movement or shadow artefacts and a vendor-specific quality index of less than 6 were excluded by the device operator.

\subsection{Spatial Variability of Graph Characteristics}
\label{app:spatial_var}
\begin{figure}[h]
    \centering
    \includegraphics[width=0.9\linewidth]{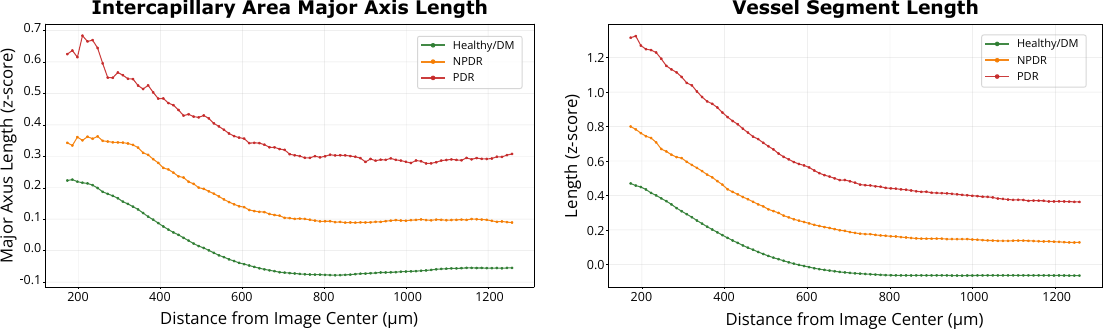}
    \caption{Dependence of major axis length of intercapillary areas and length of vessel segments on the distance from the image center. OCTA images are FAZ-centered.}
    \label{fig:feat_dist_change}
\end{figure}

Microvascular characteristics vary according to retinal location. In Figure \ref{fig:feat_dist_change}, we show how the distance from the FAZ center affects characteristics such as the vessel length and the major axis length of intercapillary areas. Characteristics are strongly influenced by disease stage and the spatial location of the vessel segment/intercapillary area within the OCTA scan.

\subsection{Interpretability Definition}
We define interpretability in our work, inspired by widely adopted definitions by Biran et al. \cite{biran2017explanation}, ``systems are interpretable if their operations can be understood by a human, either through introspection or through a produced explanation" and Miller et al \cite{miller2019explanation} ``the degree to which an observer can understand the cause of a decision". Consequently, we consider a system more interpretable if it allows a higher degree of understanding or more detailed explanations.

\subsection{Explanation Stability}
\label{app:explanation_stability}
\begin{figure}
    \centering
    \includegraphics[width=0.9\linewidth]{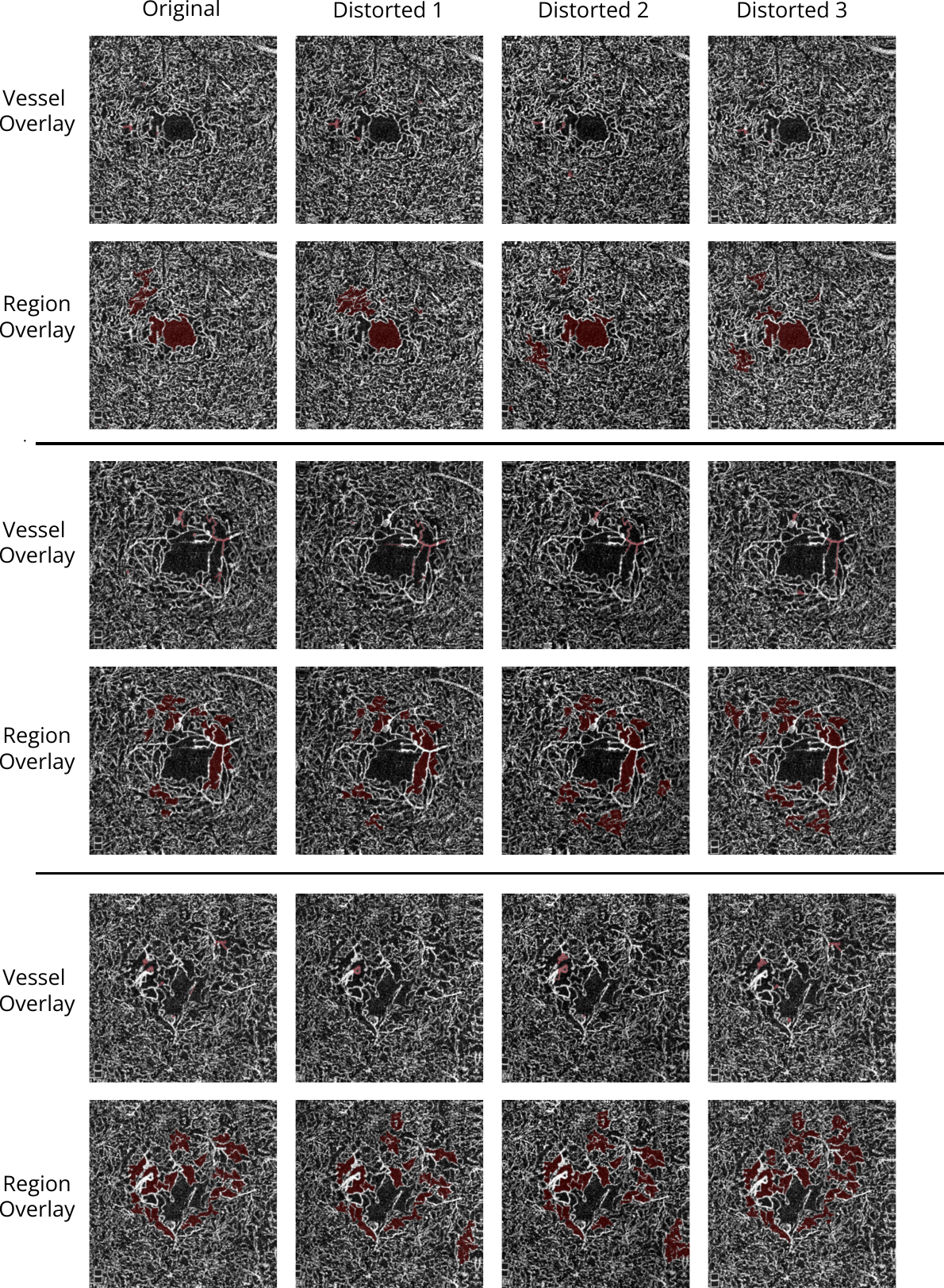}
    \caption{Spatial explanations for the most impactful vessel segments and intercapillary areas for the predicted stage label for original scans and scans from a synthetic test-retest experiment. All rows show samples with clinical label PDR, the first sample (row 1-2) is consistently classified as NPDR based on the OCTA data, while the other two samples  (row 3-6) are classified as PDR.}
    \label{fig:explanation_stability}
\end{figure}

\FloatBarrier

\subsection{Agreement of Explanations with Expert Annotations}
\label{app:expert_annotations}

\subsubsection{Metrics}\label{app:expert_annotations_metrics}

Let $\Omega$ be the pixels of an en-face scan, $M \subseteq \Omega$ the region marked as evidence for the OCTA-based diseases stage, and $A : \Omega \rightarrow \mathbb{R}_{\geq 0}$ a model's attribution map, clipped at zero and upsampled to $\Omega$ where necessary. We quantify the agreement between $A$ and $M$ with two complementary measures.

The area under the precision–recall curve (AUPRC) measures whether $A$ \emph{ranks} annotated pixels above the rest. For a threshold $t$ the selected set is $P_t = \{\, p : A(p) \geq t \,\}$, giving $\mathrm{Prec}(t) = |P_t \cap M| / |P_t|$ and $\mathrm{Rec}(t) = |P_t \cap M| / |M|$,
\begin{equation}
  \mathrm{AUPRC} = \sum_{n}
  \bigl( \mathrm{Rec}(t_n) - \mathrm{Rec}(t_{n-1}) \bigr)\,\mathrm{Prec}(t_n),
\end{equation}

where $t_1 > t_2 > \dots$ are the distinct attribution values.

The relevance mass accuracy (RMA) measures where the attribution mass is spent, irrespective of any
ranking:

\begin{equation}
  \mathrm{RMA} = \frac{\sum_{p \in M} A(p)}{\sum_{p \in \Omega} A(p)} .
\end{equation}

For a uniform map both measures reduce to the annotation prevalence $|M| / |\Omega|$, which we therefore report as the chance level. Since $\mathrm{AUPRC}$ is insensitive to how the mass is distributed and $\mathrm{RMA}$ contains no recall term, we report both.

\subsubsection{Segmentation Restricted Attribution}
For the expert annotation of abnormal vasculature, 82.3\% of the annotations overlap with the vessel segmentation. While the non-perfusion region annotation is to 91.7\% contained in the inverse of the vessel segmentation map. A restriction of the attribution to only the segmented areas for vessels and the inverse of the vessel segmentation for dropout areas, therefore, provides a structural advantage that the graph model inherently uses. We adapt the vision attribution baselines so that only attribution within the relevant areas is measured, as shown in Table \ref{tab:interpretability_combined_restricted}. As expected, this also improves attribution quality for the vision baselines, with AUPRC on abnormal vessel segments for some models even approaching that of the graph-based approach.

\input{r2_quantitative_interpretability_combined_restricted}

\FloatBarrier

\bibliographystyle{elsarticle-num} 
\bibliography{bib.bib}

\end{document}

%% file: num_params.tex
\begin{table}[t]
\centering
\caption{Comparison of model complexities and training times for 100 epochs with a fixed batch size of 16 on a single NVIDIA RTX A6000. The heterogeneous GNN is more than 100 times smaller than the smallest evaluated CNN model. The * indicates that the model is fine-tuned without adapting all learnable weights. Pretraining of the RetFound model required around two weeks with eight NVIDIA Tesla A100 \cite{zhou2023foundation}.}
\label{table:model_params}
\resizebox{\linewidth}{!}{
\begin{tabular}{c||c|c|r|r}

\multicolumn{1}{c||}{\textbf{Architecture}} 
& \multicolumn{1}{c|}{\textbf{\begin{tabular}[c]{@{}c@{}}Data\\ Representation\end{tabular}}} 
& \multicolumn{1}{c|}{\textbf{Model}} 
& \multicolumn{1}{c|}{\textbf{\begin{tabular}[c]{@{}c@{}}Model\\ Parameters\end{tabular}}} 
& \multicolumn{1}{c}{\textbf{\begin{tabular}[c]{@{}c@{}}Train\\ Time \end{tabular}}} 
\\ \hline \hline
\rule{0pt}{3ex}
\multirow{6}{*}{\textbf{CNN}}
& \multirow{6}{*}{\textbf{Image}}
& VGG11 & 133 Mio. & 13 min\\ 
& & ResNet18 & 12 Mio.  &  9 min\\
& & EfficientNet B0 & 5 Mio. & 9 min\\ 
& & ConvNeXt s & 50 Mio. & 27 min \\
& & ResNet50 & 25 Mio. & 9 min \\
& & EfficientNet V2 s & 21 Mio. & 11 min \\ [5pt]
\hline
\rule{0pt}{3ex}
\multirow{2}{*}{\textbf{Transformer}}
& \multirow{2}{*}{\textbf{Image}}
& RetFound & 87 Mio. &  *40 min
\\ 
& & Swin t & 28 Mio & 35 min \\ [5pt]
\hline
\rule{0pt}{3ex}
\textbf{\begin{tabular}[c]{@{}c@{}}Heterogeneous GNN\end{tabular}} & \textbf{Graph} & Ours & \textbf{0.04 Mio}. &  \textbf{2 min}
\\ [5pt]

\end{tabular}
}
\vspace{-0.3cm}
\end{table}

%% file: results1.tex
\begin{table*}[ht!]
\centering
\caption{Agreement scores for the staging task into fundus labels \textit{healthy}, \textit{NPDR} and \textit{PDR}. The data representation describes the information used as input to the subsequent classifier. Results are obtained on a test set that was separated before cross-validation. Standard deviations are calculated based on the results from the 5-fold cross-validation. We report p-values for paired t-tests with the heterogeneous graph model in parentheses after the standard deviations.}
\label{table:holdout}
\resizebox{\linewidth}{!}{
\begin{tabular}{c||c|c|c|c|c|c|c|c}

\multicolumn{1}{c||}{\textbf{Architecture}} 
& \multicolumn{1}{c|}{\textbf{\begin{tabular}[c]{@{}c@{}}Data\\ Representation\end{tabular}}} 
& \multicolumn{1}{c|}{\textbf{Model}} 
& \multicolumn{1}{c|}{\textbf{\begin{tabular}[c]{@{}c@{}}ROC \\ AUC \end{tabular}}}
& \multicolumn{1}{c|}{\textbf{\begin{tabular}[c]{@{}c@{}}Balanced\\ Agreement\end{tabular}}} 
& \multicolumn{1}{c|}{\textbf{\begin{tabular}[c]{@{}c@{}}F1\\ Healthy \end{tabular}}}
& \multicolumn{1}{c|}{\textbf{\begin{tabular}[c]{@{}c@{}}F1\\ NPDR \end{tabular}}}
& \multicolumn{1}{c|}{\textbf{\begin{tabular}[c]{@{}c@{}}F1 \\ PDR \end{tabular}}}
& \multicolumn{1}{c}{\textbf{\begin{tabular}[c]{@{}c@{}}DR vs no DR \\ ROC AUC \end{tabular}}}

\\ \hline\hline
\multirow{3}{*}{\textbf{Trad. ML}}
& \multirow{3}{*}{\begin{tabular}[c]{@{}c@{}}\textbf{Tabular}\\ \textbf{Biomarkers}\end{tabular}}
  & RF  & \begin{tabular}[c]{@{}c@{}} 0.796\\ \tiny$\pm$0.014(.02) \end{tabular}  & \begin{tabular}[c]{@{}c@{}} 0.639\\ \tiny$\pm$0.018(.01) \end{tabular} & \begin{tabular}[c]{@{}c@{}}0.843\\ \tiny$\pm$0.017(.09) \end{tabular} &\begin{tabular}[c]{@{}c@{}}0.421\\ \tiny$\pm$0.019(.47) \end{tabular}  & \begin{tabular}[c]{@{}c@{}}0.490\\ \tiny$\pm$0.035(.05) \end{tabular} & \begin{tabular}[c]{@{}c@{}}0.862\\ \tiny$\pm$0.007(.14) \end{tabular}  \\
& & SVM & \begin{tabular}[c]{@{}c@{}} 0.755\\ \tiny$\pm$0.023(.01) \end{tabular} & \begin{tabular}[c]{@{}c@{}} 0.633\\ \tiny$\pm$0.046(.04) \end{tabular} & \begin{tabular}[c]{@{}c@{}}0.857\\ \tiny$\pm$0.019(.01) \end{tabular} &\begin{tabular}[c]{@{}c@{}}0.341\\ \tiny$\pm$0.034(.03) \end{tabular}  & \begin{tabular}[c]{@{}c@{}}0.495\\ \tiny$\pm$0.082(.08) \end{tabular} & \begin{tabular}[c]{@{}c@{}}0.829\\ \tiny$\pm$0.027(.01) \end{tabular}  \\
\hline
\multirow{17}{*}{\textbf{CNN}}
& \multirow{12}{*}{\textbf{Image DVC}}
&      VGG11                      & \begin{tabular}[c]{@{}c@{}} 0.822\\ \tiny$\pm$0.008(.35) \end{tabular}    & \begin{tabular}[c]{@{}c@{}} 0.661\\ \tiny$\pm$0.024(.05) \end{tabular}  & \begin{tabular}[c]{@{}c@{}} 0.870\\ \tiny$\pm$0.043(.03) \end{tabular}  & \begin{tabular}[c]{@{}c@{}} 0.443\\ \tiny$\pm$0.127(.35) \end{tabular}   & \begin{tabular}[c]{@{}c@{}} 0.522\\ \tiny$\pm$0.025(.14) \end{tabular}  & \begin{tabular}[c]{@{}c@{}} 0.874\\ \tiny$\pm$0.010(.05) \end{tabular}   \\
& & ResNet18                      & \begin{tabular}[c]{@{}c@{}} 0.801\\ \tiny$\pm$0.024(.07) \end{tabular}    & \begin{tabular}[c]{@{}c@{}} 0.633\\ \tiny$\pm$0.034(.04) \end{tabular}  & \begin{tabular}[c]{@{}c@{}} 0.861\\ \tiny$\pm$0.012(.01) \end{tabular}  & \begin{tabular}[c]{@{}c@{}} 0.414\\ \tiny$\pm$0.067(.44) \end{tabular}   & \begin{tabular}[c]{@{}c@{}} 0.466\\ \tiny$\pm$0.048(.06) \end{tabular}  & \begin{tabular}[c]{@{}c@{}} 0.872\\ \tiny$\pm$0.009(.13) \end{tabular}   \\
& & Eff.Net B0                    & \begin{tabular}[c]{@{}c@{}} 0.818\\ \tiny$\pm$0.021(.30) \end{tabular}    & \begin{tabular}[c]{@{}c@{}} 0.652\\ \tiny$\pm$0.034(.10)\end{tabular}   & \begin{tabular}[c]{@{}c@{}} 0.873\\ \tiny$\pm$0.034(.01) \end{tabular}  & \begin{tabular}[c]{@{}c@{}} 0.370\\ \tiny$\pm$0.055(.17) \end{tabular}   & \begin{tabular}[c]{@{}c@{}} 0.514\\ \tiny$\pm$0.038(.17) \end{tabular}  & \begin{tabular}[c]{@{}c@{}} 0.875\\ \tiny$\pm$0.014(.12) \end{tabular}  \\
& & ResNet50                      & \begin{tabular}[c]{@{}c@{}} 0.788\\ \tiny$\pm$0.018(.02) \end{tabular}    & \begin{tabular}[c]{@{}c@{}} 0.587\\ \tiny$\pm$0.025(.01)\end{tabular}   & \begin{tabular}[c]{@{}c@{}} 0.894\\ \tiny$\pm$0.016(.01) \end{tabular}  & \begin{tabular}[c]{@{}c@{}} 0.288\\ \tiny$\pm$0.066(.03) \end{tabular}   & \begin{tabular}[c]{@{}c@{}} 0.425\\ \tiny$\pm$0.018(.02) \end{tabular}  & \begin{tabular}[c]{@{}c@{}} 0.866\\ \tiny$\pm$0.014(.43) \end{tabular}  \\
& & Eff.Net V2 s                  & \begin{tabular}[c]{@{}c@{}} 0.807\\ \tiny$\pm$0.031(.05) \end{tabular}    & \begin{tabular}[c]{@{}c@{}} 0.663\\ \tiny$\pm$0.040(.01)\end{tabular}   & \begin{tabular}[c]{@{}c@{}} 0.888\\ \tiny$\pm$0.020(.01) \end{tabular}  & \begin{tabular}[c]{@{}c@{}} 0.433\\ \tiny$\pm$0.081(.35) \end{tabular}   & \begin{tabular}[c]{@{}c@{}} 0.549\\ \tiny$\pm$0.066(.18) \end{tabular}  & \begin{tabular}[c]{@{}c@{}} 0.853\\ \tiny$\pm$0.031(.20) \end{tabular}  \\
& & ConvNeXt s                    & \begin{tabular}[c]{@{}c@{}} 0.822\\ \tiny$\pm$0.018(.37) \end{tabular}    & \begin{tabular}[c]{@{}c@{}} 0.654\\ \tiny$\pm$0.026(.05)\end{tabular}   & \begin{tabular}[c]{@{}c@{}} 0.898\\ \tiny$\pm$0.020(.01) \end{tabular}  & \begin{tabular}[c]{@{}c@{}} 0.376\\ \tiny$\pm$0.073(.22) \end{tabular}   & \begin{tabular}[c]{@{}c@{}} 0.503\\ \tiny$\pm$0.046(.10) \end{tabular}  & \begin{tabular}[c]{@{}c@{}} 0.876\\ \tiny$\pm$0.014(.11) \end{tabular}  \\ \cline{2-9}
& \textbf{Image SVC} & ConvNeXt s & \begin{tabular}[c]{@{}c@{}} 0.804\\ \tiny$\pm$0.019(.12) \end{tabular}    & \begin{tabular}[c]{@{}c@{}} 0.623\\ \tiny$\pm$0.032(.01)\end{tabular}   & \begin{tabular}[c]{@{}c@{}} 0.893\\ \tiny$\pm$0.015(.01) \end{tabular}  & \begin{tabular}[c]{@{}c@{}} 0.405\\ \tiny$\pm$0.067(.38) \end{tabular}   & \begin{tabular}[c]{@{}c@{}} 0.494\\ \tiny$\pm$0.043(.04) \end{tabular}  & \begin{tabular}[c]{@{}c@{}} 0.881\\ \tiny$\pm$0.016(.06) \end{tabular}  \\ \cline{2-9}
& \textbf{Seg. DVC} & ConvNeXt s  & \begin{tabular}[c]{@{}c@{}} 0.813\\ \tiny$\pm$0.015(.22) \end{tabular}    & \begin{tabular}[c]{@{}c@{}} 0.652\\ \tiny$\pm$0.026(.04)\end{tabular}   & \begin{tabular}[c]{@{}c@{}} 0.877\\ \tiny$\pm$0.021(.01) \end{tabular}  & \begin{tabular}[c]{@{}c@{}} 0.400\\ \tiny$\pm$0.071(.34) \end{tabular}   & \begin{tabular}[c]{@{}c@{}} 0.554\\ \tiny$\pm$0.061(.42) \end{tabular}  & \begin{tabular}[c]{@{}c@{}} 0.862\\ \tiny$\pm$0.015(.25) \end{tabular}  \\ 
 \hline

\multirow{3}{*}{\textbf{Transformer}} 
& \multirow{3}{*}{\textbf{Image}}
&  RETFound    & \begin{tabular}[c]{@{}c@{}} 0.819 \\ \tiny$\pm$0.008(.31) \end{tabular}    & \begin{tabular}[c]{@{}c@{}} 0.638\\ \tiny$\pm$0.021(.01)\end{tabular}   & \begin{tabular}[c]{@{}c@{}}0.836\\ \tiny$\pm$0.032(.16) \end{tabular}   & \begin{tabular}[c]{@{}c@{}} 0.362\\ \tiny$\pm$0.078(.13) \end{tabular}   & \begin{tabular}[c]{@{}c@{}} 0.532\\ \tiny$\pm$0.021(.14) \end{tabular}  & \begin{tabular}[c]{@{}c@{}} 0.871\\ \tiny$\pm$0.007(.13) \end{tabular}    \\
& & SwinTrans.  & \begin{tabular}[c]{@{}c@{}} 0.781 \\ \tiny$\pm$0.042(.06) \end{tabular}    & \begin{tabular}[c]{@{}c@{}} 0.556\\ \tiny$\pm$0.125(.04)\end{tabular}   & \begin{tabular}[c]{@{}c@{}}0.887\\ \tiny$\pm$0.017(.01) \end{tabular}   & \begin{tabular}[c]{@{}c@{}} 0.312\\ \tiny$\pm$0.204(.15) \end{tabular}   & \begin{tabular}[c]{@{}c@{}} 0.418\\ \tiny$\pm$0.238(.15) \end{tabular}  & \begin{tabular}[c]{@{}c@{}} 0.846\\ \tiny$\pm$0.041(.17) \end{tabular}  \\
\hline
\multirow{3}{*}{\textbf{Trad. ML}}
& \multirow{3}{*}{\begin{tabular}[c]{@{}c@{}}\textbf{Node}\\ \textbf{Embeddings}\end{tabular}} 
&     RF &\begin{tabular}[c]{@{}c@{}} 0.828\\ \tiny$\pm$0.010(.49) \end{tabular} & \begin{tabular}[c]{@{}c@{}}0.658\\ \tiny$\pm$0.007(.04)\end{tabular} & \begin{tabular}[c]{@{}c@{}}0.834\\ \tiny$\pm$0.006(.11) \end{tabular} & \begin{tabular}[c]{@{}c@{}}0.362\\ \tiny$\pm$0.031(.08) \end{tabular}  & \begin{tabular}[c]{@{}c@{}}0.560\\ \tiny$\pm$0.020(.43) \end{tabular} & \begin{tabular}[c]{@{}c@{}}0.870\\ \tiny$\pm$0.007(.19) \end{tabular}  \\
&  & SVM &\begin{tabular}[c]{@{}c@{}} 0.776\\ \tiny$\pm$0.012(.03) \end{tabular}  &  \begin{tabular}[c]{@{}c@{}}0.615\\ \tiny$\pm$0.024(.01)\end{tabular} &  \begin{tabular}[c]{@{}c@{}}0.831\\ \tiny$\pm$0.028(.07) \end{tabular} &\begin{tabular}[c]{@{}c@{}}0.395\\ \tiny$\pm$0.062(.33) \end{tabular} & \begin{tabular}[c]{@{}c@{}}0.440\\ \tiny$\pm$0.085(.04) \end{tabular} & \begin{tabular}[c]{@{}c@{}}0.867\\ \tiny$\pm$0.010(.49) \end{tabular}  \\
\hline
\textbf{Heterogeneous}
& \multirow{2}{*}{\textbf{Graph}} &  \multirow{2}{*}{Ours} & 0.829 & 0.696 & 0.819 & 0.419 & 0.568 & 0.867\\
\textbf{Graph Learning} &  & & \tiny$ \pm$0.037  & \tiny$\pm$0.034 & \tiny$\pm$0.021 & \tiny$\pm$0.057 & \tiny$\pm$0.090 & \tiny$\pm$0.003 \\
\end{tabular}
}
\end{table*}

%% file: results2.tex
\begin{table}[t]
\centering
\caption{Classification performance for the binary classification into \textit{healthy} versus DR. The results are obtained on the external OCTA-500 without further fine-tuning the models previously trained on the other reported dataset. The checkpoint from the 5-fold cross-validation on the proprietary dataset with the highest balanced agreement is used for evaluation on the OCTA-500 dataset.}
\label{table:octa500}
\resizebox{\linewidth}{!}{
\begin{tabular}{c||c|c|c|c|c|c|c}
\multicolumn{1}{c||}{\textbf{Architecture}} & \multicolumn{1}{c|}{\textbf{\begin{tabular}[c]{@{}c@{}}Data\\ Rep.\end{tabular}}} & \multicolumn{1}{c|}{\textbf{Model}} & \multicolumn{1}{c|}{\textbf{\begin{tabular}[c]{@{}c@{}}Bal.\\ Agr.\end{tabular}}}
& \multicolumn{1}{c|}{\textbf{\begin{tabular}[c]{@{}c@{}}F1\\ Heal. \end{tabular}}}
& \multicolumn{1}{c|}{\textbf{\begin{tabular}[c]{@{}c@{}}F1\\ DR \end{tabular}}}
& \multicolumn{1}{c|}{\textbf{\begin{tabular}[c]{@{}c@{}}AUC\\ ROC \end{tabular}}}
& \multicolumn{1}{c}{\textbf{\begin{tabular}[c]{@{}c@{}}AUC\\ PRC \end{tabular}}}
\\ \hline\hline

\multirow{2}{*}{\textbf{Trad. ML}}
& \multirow{2}{*}{\begin{tabular}[c]{@{}c@{}}\textbf{Tabular}\\ \textbf{Biom.}\end{tabular}}
 & RF & 0.821 & 0.963 & 0.760 & 0.960 & 0.872 \\
& & SVM & 0.806 & 0.947 & 0.691 & 0.876 & 0.773\\
\hline
\multirow{3}{*}{\textbf{CNN}}
& \multirow{3}{*}{\textbf{Image}}
& 
VGG11 & 0.773& 0.958 & 0.696 & 0.942 & 0.873
\\
& &
ResNet18 & 0.586 & 0.930 & 0.294 & 0.920 & 0.785\\
& & Eff.Net B0 & 0.707 & 0.950 & 0.585 & 0.862 & 0.778 \\
 \hline
\multirow{1}{*}{\textbf{Transformer}} & \textbf{Image} & RETFound & 0.746 & 0.915 & 0.558 & 0.777 & 0.628\\
\hline
\multirow{2}{*}{\textbf{Trad. ML}}
& \textbf{Node} & RF & 0.839 & 0.967 & 0.785 & 0.916 & 0.857\\
& \textbf{Embed.} & SVM & 0.500 & 0.000 & 0.266 & 0.289 & 0.111 \\
\hline
\textbf{Heterogeneous}
& \multirow{2}{*}{\textbf{Graph}} &  \multirow{2}{*}{Ours} &  \multirow{2}{*}{0.893} & \multirow{2}{*}{0.978} &  \multirow{2}{*}{0.868} & \multirow{2}{*}{0.952} &  \multirow{2}{*}{0.909}\\
\textbf{Graph Learning} & & & & & & &\\
\end{tabular}
}
\end{table}

%% file: results3.tex
\begin{table*}[h]
    \centering
    \caption{Comparison for using vessel graph, intercapillary area graph, and heterogeneous graph}
    \label{table:homog}
    \resizebox{\linewidth}{!}{
    \begin{tabular}{c||c|c|c|c|c|c|c}
    
    \multicolumn{1}{c|}{\textbf{\begin{tabular}[c]{@{}c@{}}Data\\ Representation\end{tabular}}} 
    & \multicolumn{1}{c|}{\textbf{Model}} 
    & \multicolumn{1}{c|}{\textbf{\begin{tabular}[c]{@{}c@{}}ROC \\ AUC \end{tabular}}}
    & \multicolumn{1}{c|}{\textbf{\begin{tabular}[c]{@{}c@{}}Bal.\\ Agr.\end{tabular}}} 
    & \multicolumn{1}{c|}{\textbf{\begin{tabular}[c]{@{}c@{}}F1\\ Healthy \end{tabular}}}
    & \multicolumn{1}{c|}{\textbf{\begin{tabular}[c]{@{}c@{}}F1\\ NPDR \end{tabular}}}
    & \multicolumn{1}{c|}{\textbf{\begin{tabular}[c]{@{}c@{}}F1 \\ PDR \end{tabular}}}
    & \multicolumn{1}{c}{\textbf{\begin{tabular}[c]{@{}c@{}}DR vs no DR \\ROC AUC \end{tabular}}}
    
    \\ \hline\hline
    \textbf{Vessel Graph} &  GNN                                                              & \begin{tabular}[c]{@{}c@{}} 0.828\\ \scriptsize$\pm$0.005 \end{tabular} & \begin{tabular}[c]{@{}c@{}} 0.642\\ \scriptsize$\pm$0.025 \end{tabular}  & \begin{tabular}[c]{@{}c@{}} 0.812\\ \scriptsize$\pm$0.037 \end{tabular}  & \begin{tabular}[c]{@{}c@{}} 0.372\\ \scriptsize$\pm$0.060 \end{tabular}   & \begin{tabular}[c]{@{}c@{}} 0.516\\ \scriptsize$\pm$0.048 \end{tabular}  & \begin{tabular}[c]{@{}c@{}} 0.876\\ \scriptsize$\pm$0.010 \end{tabular}  \\
    \textbf{\begin{tabular}[c]{@{}c@{}}Intercapillary \\ Area Graph \end{tabular}} &  GNN     & \begin{tabular}[c]{@{}c@{}} 0.815\\ \scriptsize$\pm$0.018 \end{tabular} & \begin{tabular}[c]{@{}c@{}} 0.627\\ \scriptsize$\pm$0.043 \end{tabular}  & \begin{tabular}[c]{@{}c@{}} 0.836\\ \scriptsize$\pm$0.047 \end{tabular}  & \begin{tabular}[c]{@{}c@{}} 0.336\\ \scriptsize$\pm$0.178 \end{tabular}   & \begin{tabular}[c]{@{}c@{}} 0.498\\ \scriptsize$\pm$0.068 \end{tabular}  & \begin{tabular}[c]{@{}c@{}} 0.881\\ \scriptsize$\pm$0.012 \end{tabular}  \\
    \textbf{Heterogeneous Graph}                                                   &  GNN     & \begin{tabular}[c]{@{}c@{}} 0.829\\ \scriptsize$\pm$0.037 \end{tabular} & \begin{tabular}[c]{@{}c@{}} 0.696\\ \scriptsize$\pm$0.034 \end{tabular}  & \begin{tabular}[c]{@{}c@{}} 0.819\\ \scriptsize$\pm$0.021 \end{tabular}  & \begin{tabular}[c]{@{}c@{}} 0.419\\ \scriptsize$\pm$0.057 \end{tabular}   & \begin{tabular}[c]{@{}c@{}} 0.568\\ \scriptsize$\pm$0.090 \end{tabular}  & \begin{tabular}[c]{@{}c@{}} 0.867\\ \scriptsize$\pm$0.003 \end{tabular}  \\ 
     \hline
    \end{tabular}
    }
    \end{table*}

%% file: r2_quantitative_interpretability_combined.tex
\begin{table}[h]
\centering
\caption{Quantitative agreement of model explanations with expert annotations of abnormal vessels and capillary dropout, $n=11$ eyes (4 NPDR, 7 PDR) from held-out data, scored on the $1216\times1216$ grid.}
\label{tab:interpretability_combined}
\small
\setlength{\tabcolsep}{4pt}
\begin{tabular}{ll|cc|cc}
& & \multicolumn{2}{c|}{\textbf{Abnormal vessels}} & \multicolumn{2}{c}{\textbf{Capillary dropout}} \\
\textbf{Model} & \textbf{Attr.} & AUPRC & RMA & AUPRC & RMA \\
\hline
\textbf{Graph (Sec. \ref{sec:methods})} & IG (Sec. \ref{sec:methods}) & \textbf{0.175} & \textbf{0.306} & \textbf{0.498} & \textbf{0.481} \\
\hline
ConvNeXt s & IG & 0.023 & 0.018 & 0.297 & 0.197 \\
 & Grad-CAM & 0.049 & 0.025 & 0.156 & 0.102 \\
Eff.Net b0 & IG & 0.019 & 0.010 & 0.055 & 0.043 \\
 & Grad-CAM & 0.032 & 0.013 & 0.087 & 0.050 \\
EffNet-V2-S & IG & 0.017 & 0.010 & 0.055 & 0.045 \\
 & Grad-CAM & 0.029 & 0.011 & 0.099 & 0.045 \\
ResNet18 & IG & 0.011 & 0.009 & 0.080 & 0.054 \\
 & Grad-CAM & 0.025 & 0.012 & 0.068 & 0.049 \\
ResNet50 & IG & 0.020 & 0.015 & 0.264 & 0.148 \\
 & Grad-CAM & 0.013 & 0.011 & 0.046 & 0.036 \\
VGG-11 & IG & 0.030 & 0.021 & 0.190 & 0.110 \\
 & Grad-CAM & 0.015 & 0.012 & 0.094 & 0.065 \\
\hline
\textit{chance} & & 0.007 & 0.007 & 0.030 & 0.030 \\
\hline
\end{tabular}
\end{table}

%% file: stability_segmentation.tex
\begin{table}[h]
\centering
\textcolor{black}{
\caption{Dice scores between segmentations of a synthetic test-retest scenario with five distorted raw images per original input image.}
\label{tab:dice_foreground_combined}
\begin{tabular}{l|c|c|c}
\textbf{Stage} & \textbf{Vessel Dice} & \textbf{ICA Dice} & \textbf{FG Ratio} \\ \hline
No DR & $0.9675 \pm 0.0085$ & $0.9722 \pm 0.0069$ & $0.4666 \pm 0.0226$\\
NPDR & $0.9643 \pm 0.0103$ & $0.9740 \pm 0.0069$ & $0.4316 \pm 0.0313$ \\
PDR & $0.9610 \pm 0.0121$ & $0.9756 \pm 0.0069$ & $0.4000 \pm 0.0434$ \\
\hline \hline
All & $0.9665 \pm 0.0094$ & $0.9727 \pm 0.0070$ & $0.4561 \pm 0.0331$ \\
\end{tabular}
}
\end{table}

%% file: stability_graphs_v2.tex
\begin{table}[ht]
\centering
\textcolor{black}{
\caption{Node count analysis for vessel segments (VS) and intercapillary areas (ICA) for the synthetic test-retest stability analysis. The count column shows the average number of nodes stratified by disease class with absolute and relative standard deviations. The last column shows the standard deviation for the node count for the distorted and original sample of the test-retest (T-R) scenario.}
\label{tab:combined_graph_ica_2D_TTA}
\begin{tabular}{l|l|c|c}
\textbf{Stage} & \textbf{Node} & \textbf{Count $\pm $ SD  Abs./Rel.(\%)} & \textbf{T-R SD (\%)} \\ \hline
\multirow{2}{*}{No DR} & VS & $4653.5 \pm 467.7$ / $10.05$ & $1.80$ \\
                       & ICA        & $1482.3 \pm 185.4$ / $12.51$ & $2.30$ \\ \hline
\multirow{2}{*}{NPDR} & VS & $3917.3 \pm 572.0$ / $14.60$ & $2.80$ \\
                       & ICA        & $1186.4 \pm 226.9$ / $19.13$ & $3.54$ \\ \hline
\multirow{2}{*}{PDR} & VS & $3363.4 \pm 678.7$ / $20.18$ & $4.15$ \\
                       & ICA        & $964.9 \pm 259.0$ / $26.84$ & $5.08$ \\ \hline \hline
\multirow{2}{*}{All}   & VS & $4441.6 \pm 646.4$ / $14.55\%$ & $2.13$ \\
                       & ICA       & $1397.2 \pm 256.8$ / $18.38\%$ & $2.70$ \\
\end{tabular}
}
\end{table}

%% file: stability_classification.tex
\begin{table}[h]
\centering
\textcolor{black}{
\caption{Prediction stability across original and augmented samples, stratified by disease stage (5-fold cross-validation validation scores). The agreement rate measures the fraction of predictions matching the majority vote. The consistency rate measures the fraction of augmented predictions matching the original sample's prediction.}
\label{tab:stability_validation}
\begin{tabular}{l|c|c}
\textbf{Stage} & \textbf{Agreement Rate} & \textbf{Consistency Rate} \\ \hline
No DR & $94.0\% \pm 1.7\%$ & $91.6\% \pm 2.3\%$ \\
NPDR       & $88.7\% \pm 2.1\%$ & $82.9\% \pm 3.0\%$ \\
PDR        & $93.5\% \pm 2.7\%$ & $90.6\% \pm 5.1\%$ \\ \hline
All        & $93.4\% \pm 1.5\%$ & $90.6\% \pm 2.3\%$ \\
\end{tabular}
}
\end{table}

%% file: stage_deviations.tex
\begin{table}[h]
\centering
\textcolor{black}{
\caption{Stage deviation distribution between original and augmented predictions (5-fold cross-validation validation scores). Each row shows the fraction of augmented predictions that deviate by 0, 1, or 2 disease stages from the predicted stage for the original sample.}
\label{tab:stage_deviation_validation}
\begin{tabular}{l|c|c|c}
\textbf{GT Stage} & \textbf{0-stages} & \textbf{1-stage} & \textbf{2-stages} \\ \hline
No DR   & $91.6\% \pm 2.3\%$ & $8.1\% \pm 2.3\%$ & $0.3\% \pm 0.4\%$ \\
NPDR         & $82.9\% \pm 3.0\%$ & $17.0\% \pm 3.3\%$ & $0.2\% \pm 0.4\%$ \\
PDR          & $90.6\% \pm 5.1\%$ & $7.6\% \pm 3.4\%$ & $1.9\% \pm 2.0\%$ \\
\hline
All          & $90.6\% \pm 2.3\%$ & $9.0\% \pm 2.2\%$ & $0.4\% \pm 0.3\%$ \\
\end{tabular}
}
\end{table}

%% file: stability_distortion_parameters.tex
\begin{table}[h]
\centering
\renewcommand{\arraystretch}{1.3}
\caption{Distortion parameters for the simulation of intra-patient scan variability in a test-retest scenario.}
\label{tab:distortion_params}
\begin{tabular}{l|l|c}
\hline
\textbf{Distortion Type} & \textbf{Parameter} & \textbf{Range} \\ \hline
Rigid Transformation & Translation X & $[-24.72, 16.92]$ pix. \\
 & Translation Y & $[-23.20, 31.24]$ pix. \\
 & Rotation & $[-5.00, 1.74]$ deg. \\ \hline
Spatial Gradient & Strength & $[0.052, 0.114]$ \\
 & Angle & $[9.3, 297.0]$ deg. \\ \hline
Gaussian Noise & Std & $[0.315, 0.956]$ \\ \hline
Brightness/Contrast & Brightness offset & $[-3.52, 4.67]$ \\
 & Contrast multiplier & $[0.966, 1.043]$ \\ \hline
Signal Strength & Gamma factor & $[0.922, 1.068]$ \\ \hline
Signal Attenuation & Fraction of samples  & $20\%$ of samples \\ 
 & Region side length&  $[40, 119]$ pix. \\ 
 & Attenuation factor  & $[0.3, 0.7]$ \\ \hline
\end{tabular}
\end{table}

%% file: r2_cohort_demographics.tex
\begin{table}[h]
\centering
\renewcommand{\arraystretch}{1.3}
\caption{Demographic characteristics of the main clinical cohort, stratified
into the no DR class (healthy and DM without DR) and the DR class (NPDR and
PDR) as defined in Section~3.4.1, and of the test-retest cohort. Demographics
are reported per patient; for 419 of the 849 patients of the main cohort both
eyes were imaged, yielding the 1268 images reported in Section~3.4.1. The age
SD is reported where individual-level data were available.}
\label{tab:cohort_demographics}
\begin{tabular}{l|c|c|c}
\hline
\textbf{Cohort} & \textbf{n} & \textbf{Age (y)} & \textbf{F / M} \\ \hline
Main cohort --- no DR & 640 & 49.3 & 327 / 313 \\
Main cohort --- DR & 209 & 56.2 & 107 / 102 \\
Test-retest & 15 & $30.3 \pm 6.8$ & 9 / 6 \\ \hline
\end{tabular}
\end{table}

%% file: r2_quantitative_interpretability_combined_restricted.tex
\begin{table}[ht]
\centering
\caption{Quantitative agreement of model explanations with expert annotations of abnormal vessels and capillary dropout, $n=11$ eyes (4 NPDR, 7 PDR) from held-out data, scored on the $1216\times1216$ grid. Here, every attribution map is first masked to the segmented vasculature or its complement, giving the vision baselines the same structural prior that the graph model has by construction. The graph model is therefore unchanged from Table~\ref{tab:interpretability_combined}.}
\label{tab:interpretability_combined_restricted}
\small
\setlength{\tabcolsep}{4pt}
\begin{tabular}{ll|cc|cc}
& & \multicolumn{2}{c|}{\textbf{Abnormal vessels}} & \multicolumn{2}{c}{\textbf{Capillary dropout}} \\
\textbf{Model} & \textbf{Attr.} & AUPRC & RMA & AUPRC & RMA \\
\hline
\textbf{Graph (Sec. \ref{sec:methods})} & IG (Sec. \ref{sec:methods}) & \textbf{0.175} & \textbf{0.306} & \textbf{0.498} & \textbf{0.481} \\
\hline
ConvNeXt s & IG & 0.102 & 0.083 & 0.304 & 0.231 \\
 & Grad-CAM & 0.097 & 0.063 & 0.199 & 0.147 \\
Eff.Net b0 & IG & 0.068 & 0.030 & 0.065 & 0.057 \\
 & Grad-CAM & 0.062 & 0.026 & 0.111 & 0.075 \\
EffNet v2 s & IG & 0.066 & 0.031 & 0.062 & 0.057 \\
 & Grad-CAM & 0.060 & 0.022 & 0.127 & 0.068 \\
ResNet18 & IG & 0.055 & 0.028 & 0.085 & 0.069 \\
 & Grad-CAM & 0.076 & 0.030 & 0.085 & 0.071 \\
ResNet50 & IG & 0.085 & 0.051 & 0.279 & 0.186 \\
 & Grad-CAM & 0.023 & 0.021 & 0.067 & 0.054 \\
VGG11 & IG & 0.162 & 0.081 & 0.203 & 0.137 \\
 & Grad-CAM & 0.026 & 0.027 & 0.113 & 0.091 \\
\hline
\textit{chance, restricted} & & 0.013 & 0.014 & 0.045 & 0.046 \\
\textit{max attainable} & & 0.823 & 1.000 & 0.917 & 1.000 \\
\hline
\end{tabular}
\end{table}